\documentclass[pdflatex,sn-mathphys-num]{sn-jnl}% Math and Physical Sciences Numbered Reference Style
\usepackage{graphicx}%
\usepackage{multirow}%
\usepackage{amsmath,amssymb,amsfonts}%
\usepackage{amsthm}%
\usepackage{mathrsfs}%
\usepackage[title]{appendix}%
\usepackage{xcolor}%
\usepackage{textcomp}%
\usepackage{manyfoot}%
\usepackage{booktabs}%
\usepackage{algorithm}%
\usepackage{algorithmicx}%
\usepackage{algpseudocode}%
\usepackage{listings}%
\usepackage[utf8]{inputenc} % allow utf-8 input
\usepackage[T1]{fontenc}    % use 8-bit T1 fonts
\usepackage{hyperref}       % hyperlinks
\usepackage{url}            % simple URL typesetting
\usepackage{nicefrac}       % compact symbols for 1/2, etc.
\usepackage{microtype}      % microtypography
\usepackage{doi}
\usepackage{subcaption}

\usepackage{amsmath,amsfonts,amssymb}
\usepackage{graphicx}
\usepackage{booktabs}
\usepackage{physics}
\usepackage{siunitx}
\usepackage{bm}
\usepackage{color}
\usepackage{algorithm}
\usepackage{algpseudocode}
\usepackage{geometry}

\usepackage{titlesec}
\titlespacing*{\section}{0pt}{1ex}{1ex}
\titlespacing*{\subsection}{0pt}{0.6ex}{0.6ex}
\titlespacing*{\subsubsection}{0pt}{0.4ex}{0.4ex}

\usepackage{enumitem}

\titleformat{\section}
  {\normalfont\fontsize{11}{11}\bfseries}{\thesection}{1em}{}
  \titleformat{\subsection}
  {\normalfont\fontsize{11}{11}\bfseries}{\thesubsection}{1em}{}

  \titleformat{\subsubsection}
  {\normalfont\fontsize{11}{11}\bfseries}{\thesubsubsection}{1em}{}

\usepackage{tikz}
\usetikzlibrary{shapes,calc,positioning}
\usetikzlibrary{decorations.pathmorphing,arrows}
\usetikzlibrary{arrows.meta,fit}
\usepackage{tcolorbox}
\usepackage{soul}

\tikzset{
        block_small/.style = {draw, rectangle,
            minimum height=1cm,
            minimum width=1cm},
        block/.style = {draw, rectangle,
            minimum height=1cm,
            minimum width=2cm},
        input/.style = {coordinate,node distance=1cm},
        output/.style = {coordinate,node distance=4cm},
        arrow/.style={draw, -latex,node distance=2cm},
        pinstyle/.style = {pin edge={latex-, black,node distance=2cm}},
        sum/.style = {draw, circle, node distance=1cm},
}

\theoremstyle{thmstyleone}%
\theoremstyle{thmstyletwo}%
\newtheorem{remark}{Remark}%

\theoremstyle{thmstylethree}%

\newcommand{\vecf}{\mathbf{f}}
\newcommand{\vecx}{\mathbf{x}}
\newcommand{\vecu}{\mathbf{u}}
\newcommand{\vecv}{\mathbf{v}}
\newcommand{\vecy}{\mathbf{y}}
\newcommand{\vecz}{\mathbf{z}}

\newcommand{\reviewerone}[1]{\textcolor{black}{#1}}
\newcommand{\reviewertwo}[1]{\textcolor{black}{#1}}
\newcommand{\reviewerthree}[1]{\textcolor{black}{#1}}

\begin{document}

\title[Toward Adaptive Non-Intrusive Reduced-Order Models: Design and Challenges]{\reviewerone{Toward Adaptive Non-Intrusive Reduced-Order Models: Design and Challenges}}

\author*[1]{\fnm{Amirpasha} \sur{Hedayat}}\email{ahedayat@umich.edu}

\author[2]{\fnm{Alberto} \sur{Padovan}}\email{alberto.padovan@njit.edu}

\author[1]{\fnm{Karthik} \sur{Duraisamy}}\email{kdur@umich.edu}

\affil*[1]{\orgdiv{Department of Aerospace Engineering}, \orgname{University of Michigan}, \orgaddress{\city{Ann Arbor}, \postcode{48105}, \state{MI}, \country{USA}}}

\affil[2]{\orgdiv{Department of Mechanical and Industrial Engineering}, \orgname{New Jersey Institute of Technology}, \orgaddress{\city{Newark}, \postcode{07102}, \state{NJ}, \country{USA}}}

% \affil[3]{\orgdiv{Department}, \orgname{Organization}, \orgaddress{\street{Street}, \city{City}, \postcode{610101}, \state{State}, \country{Country}}}

\abstract{Projection-based Reduced Order Models (ROMs) are often deployed as static surrogates, which limits their practical utility once a system leaves the training manifold. We formalize and study adaptive non-intrusive ROMs that update both the latent subspace and the reduced dynamics online. Building on ideas from static non-intrusive ROMs, specifically, Operator Inference (OpInf) and the recently-introduced Non-intrusive Trajectory-based optimization of Reduced-Order Models (NiTROM), we propose three formulations: Adaptive OpInf (sequential basis/operator refits), Adaptive NiTROM (joint Riemannian optimization of encoder/decoder and polynomial dynamics), and a hybrid that initializes NiTROM with an OpInf update.
We describe the online data window, adaptation window, and computational budget, and analyze cost scaling. On a transiently perturbed lid‑driven cavity flow, static Galerkin/OpInf/NiTROM drift or destabilize when forecasting beyond training. In contrast, Adaptive OpInf robustly suppresses amplitude drift with modest cost; Adaptive NiTROM is shown to attain near‑exact energy tracking under frequent updates but is sensitive to its initialization and optimization depth; the hybrid is most reliable under regime changes and minimal offline data, yielding physically coherent fields and bounded energy. We argue that predictive claims for ROMs must be cost‑aware and transparent, with clear separation of training/adaptation/deployment regimes and explicit reporting of online budgets and full-order model queries. This work provides a practical \reviewerone{template} for building self‑correcting, non‑intrusive ROMs that remain effective as the dynamics evolve well beyond the initial manifold.}


\maketitle

%%%%%%%%%%%%%%%%%%%%%%%%%%%%%%%%%%%%%%%%%%%%%%%%%%%%%%%%%%%%%
\section{Introduction}
\label{sec:intro}
%%%%%%%%%%%%%%%%%%%%%%%%%%%%%%%%%%%%%%%%%%%%%%%%%%%%%%%%%%%%%

The cost of repeatedly solving large systems of  differential equations often limits the applicability of numerical simulations in several engineering  tasks including optimization, uncertainty quantification, control, and digital twins.
This problem can be addressed by replacing (or complementing) the expensive high-fidelity solver with a  reduced-order model (ROM) that can be solved at a fraction of the original computational cost~\citep{benner2015survey,antoulas2020interpolatory,rowley2017model}.
Simply put, the field of model reduction concerns itself with the development of models that can maximize computational gains while minimizing loss of accuracy. While ROMs have garnered increasing attention in the research community, their adoption in practical engineering applications remains limited.

In this work, we approach the problem of \emph{non-intrusive} projection-based reduced-order modeling to accelerate costly simulations of physical systems. The goals of this work are two-fold: first, to characterize the challenges in achieving predictive capabilities with current ROM approaches; and second, to recommend ways forward through adaptive non-intrusive formulations. Specifically, we explore the advantages and drawbacks of formulations based on least-squares regression, as well as a novel paradigm based on a \emph{joint}, \emph{dynamics-aware} adaptation of the trial and test subspaces and latent-space dynamics.
These formulations build upon Operator Inference~\cite{peherstorfer2016opinf}, and the recently-introduced Non-intrusive Trajectory-based optimization of Reduced-Order Models (NiTROM)~\cite{padovan2024}.
\reviewerone{The primary contribution of this work is therefore the development and systematic evaluation of a
practical online adaptation workflow for existing non-intrusive ROMs, together with a novel model
that combines the ideas of Operator Inference and NiTROM.}
In the rest of the section we provide a brief review of existing static and adaptive, intrusive and non-intrusive reduced-order modeling strategies.

Given a high-fidelity model with state dimension $n \gg 1$, static and adaptive projection-based ROMs are typically obtained by constraining the state to evolve on a low-dimensional subspace with dimension $r \ll n$. Since the adaptive ROM utilizes online information, it generally operates on a lower-dimensional subspace than the static ROM.
This subspace, and the latent-space coordinates on it, are implicitly defined by the choice of so-called decoders and encoders, respectively, which are maps from the $r$-dimensional space to the ambient $n$-dimensional space, and vice versa.
The evolution of the state on the target subspace is governed by a $r$-dimensional dynamical system, which is much cheaper to integrate than the original high-fidelity solver. Adaptive ROMs pursue interface with the high-fidelity operator to update the latent space.
These building blocks (i.e., the encoder, decoder, latent-space dynamics and manifold updates) can be computed using intrusive, or non-intrusive algorithms, as discussed below.

\subsection{Intrusive reduced-order models}

An intrusive ROM is one where the computation of the encoder, decoder, and latent-space dynamics requires  direct access to the underlying high-fidelity operator. 
In most projection-based ROMs, the encoder and decoder are linear maps defining a rank-$r$ projection onto the span of the decoder, and the latent-space dynamics are obtained by projecting the full-order dynamics.
Classical examples are Galerkin models obtained by \emph{orthogonal} projection onto Proper Orthogonal Decomposition (POD) modes~\citep{lumley1967structure, sirovich1987turbulence, holmes1996turbulence} associated with a representative training data set~\citep{Rowley2004model,barone2009}.
This procedure is ``weakly'' intrusive, since the computation of the encoder and decoder is purely data-driven, while the computation of the latent-space dynamics requires access only to the high-fidelity dynamics.
Despite the widespread use of these models, it is well-known that orthogonally projecting the system onto high-energy/high-variance subspaces can cause the accidental truncation of low-energy/low-variance features that are nonetheless dynamically important~\citep{chomaz2005,rowley2017model,ahmed2021closures,rezaian2022non}.
This can ultimately lead to models with poor predictive accuracy, especially along transient trajectories.

Petrov-Galerkin models constructed using appropriately-defined \emph{oblique} projections can address this shortcoming. 
Notable examples of ROM frameworks performing such oblique projections include the following. 
The Least-Squares Petrov–Galerkin (LSPG) method~\citep{carlberg2011, carlberg2017} builds a reduced model by minimizing the residual of the discrete governing equations in a least-squares sense, leading to stable time-discrete ROMs. 
Another prominent approach is Balanced Truncation (BT)~\citep{moore1981, willcox2002}, which constructs a ROM by balancing the controllability and observability Gramians of the linearized system, thereby retaining the most dynamically important modes.
Moreover, Balanced POD (BPOD)~\citep{rowley2005} provides an approximate version of BT that uses snapshots of primal and adjoint simulations to compute empirical Gramians.

More recent formulations include Trajectory-based Optimization of Oblique Projections (TrOOP)~\citep{otto2022}, which poses the ROM construction as an optimization problem seeking the best oblique projection to minimize the forecasting error along training trajectories, and Covariance Balancing Reduction using Adjoint Snapshots (CoBRAS)~\citep{otto2023,Zanardi2025}, which balances covariance operators derived from primal and adjoint snapshot ensembles to achieve robust and balanced ROMs.
These methods are highly intrusive since model computation and assembly requires access to the nonlinear dynamics, as well as to the linearized adjoint of the latter around (possibly time-varying) solutions of the full-order model.
This hinders the applicability of these formulations to high-dimensional systems simulated using a black-box solver (e.g., commercial and legacy codes), where access to the source code is limited for proprietary reasons.

\subsection{Non-intrusive reduced-order models}

Non-intrusive formulations seek to address the aforementioned problem by developing algorithms that rely exclusively on simulation (or experimental) data, without requiring access to the source code of the high-fidelity solver. 
These algorithms are therefore broadly applicable to systems that are simulated using commercial and legacy codes.
Early examples include data-driven modal-decomposition-based methods such as Dynamic Mode Decomposition (DMD)~\citep{schmid2010,rowley2009}, as well as more recent extensions that employ machine-learning architectures and operator learning~\citep{Hesthaven2018,Bhattacharya2021,xu2020}. Further, Balanced Truncation can also be achieved in a non-intrusive setting~\cite{rezaian2022non,rezaian2023data}.
Among the most widely adopted non-intrusive approaches is Operator Inference (OpInf), which recasts the reduced model identification as a least-squares regression problem that learns polynomial operators directly in the reduced space~\citep{peherstorfer2016opinf,kramer2024opinf}. 

OpInf has been  applied to complex nonlinear dynamical systems~\citep{qian2020,qian2019,mcquarrie2021,swischuk2020,benner2020} and parametric problems~\citep{mcquarrie2023}, and has also been extended to quadratic manifolds~\citep{geelen2023}. 
However, this framework seeks models in the span of an orthogonal projection operator defined in terms of high-energy/high-variance POD modes, and this can ultimately lead to models with poor forecasting accuracy. 
The recently proposed Non-intrusive Trajectory-based optimization of Reduced-Order Models (NiTROM) framework~\citep{padovan2024} provides a non-intrusive learning setting that seeks to address this drawback. 
NiTROM jointly optimizes oblique-projection operators and reduced-order polynomial dynamics against available high-fidelity trajectory data. This approach has been shown to offer significant improvements in model forecasting accuracy, especially in systems exhibiting high sensitivity to low-energy features and large-amplitude transient growth (e.g., advection-dominated and high-shear flows).

\subsection{Adaptive model reduction}

A number of  researchers have explored adaptive ROMs within the intrusive setting. 
In such paradigms, adaptation typically involves updating the reduced basis and the hyper-reduction sampling locations, while the reduced operators are automatically refreshed through their dependence on the FOM operators. 
Early works by Amsallem and Farhat~\cite{amsallem2008,amsallem2011,amsallem2012} pioneered interpolation-based ROMs that adapt across parameters by constructing local reduced bases and interpolating operators on matrix manifolds. 
These approaches tailor a group of reduced subspaces to different regions of the parameter domain, rather than relying on a single global subspace. 
Subsequent methods, such as the ``Localized Discrete Empirical Interpolation Method'' (LDEIM) of Peherstorfer et al.~\cite{peherstorfer2014ldeim}, extended this idea to the hyper-reduction stage by clustering snapshots and constructing multiple local DEIM spaces and sampling points, enabling efficient nonlinear model evaluations across distinct dynamical regimes.
Beyond these, Peherstorfer and collaborators advanced the field toward \emph{truly} online adaptive frameworks. 
In their ``Dynamic Data-Driven ROMs''~\cite{peherstorfer2015dddrom}, they proposed performing low-rank updates to the reduced operators directly from sensor data.
This direction was further developed through the Adaptive DEIM (ADEIM)~\cite{peherstprfer2015adeim} and its successor, AADEIM~\cite{peherstorfer2020aadeim}, which perform on-the-fly, low-rank updates to both the reduced basis and the interpolation points based on localized residual information.

Other strategies, such as the lookahead sampling technique of Singh et al.~\cite{singh2023}, predict near-future system states to guide data collection and update scheduling. 
From a geometric optimization viewpoint, Zimmermann et al.~\cite{zimmermann2018} proposed employing the ``Grassmannian Rank-One Update Subspace Estimation'' (GROUSE) method~\cite{balzano2010,balzano2013}, which performs continuous rank-one updates to the reduced basis along geodesics on the Grassmann manifold, yielding closed-form subspace adaptation formulas.
Similarly, Huang and Duraisamy~\cite{huang2023} introduced a rank-one adaptive ROM formulation that incrementally corrects the subspace to enforce zero projection error for each new system snapshot. Their results demonstrated that this efficient one-step update preserves long-term stability and physical fidelity, even in chaotic reacting-flow systems. Recently, Mohaghegh and Huang~\cite{mohaghegh2026} developed feature-guided sampling that introduces physics-informed criteria to direct online sampling toward dynamically active regions (in their cases, shocks and flame fronts), further enhancing adaptivity in hyper-reduction. Beyond these, many other efforts have explored adaptive ROMs, including hybrid snapshot and FOM/ROM coupling approaches~\cite{zucatti2024,bai2022,feng2021}, adaptive model-reduction methods for optimization~\cite{yano2021}, basis refinement~\cite{carlberg2015href,etter2020}, and time-dependent basis frameworks~\cite{patil2023,jung2025,ramezanian2021}.
Despite these advances, a key limitation of all such methods is their reliance on intrusive access to the FOM.
Specifically, they require explicit evaluation of residuals, Jacobians, or projection operators to perform sampling and basis updates.
In non-intrusive settings, where only state and input data are available, such information is not accessible, and any parameter update must be performed in a purely data-driven sense.
Another source of complexity arises from the fact that in an adaptive non-intrusive ROM, one must not only update the reduced basis but also re-identify the reduced operators themselves, which are learned directly from data rather than obtained through projection.

Some recent efforts have explored limited forms of adaptivity within non-intrusive ROMs. 
For instance, Geelen and Willcox~\cite{geelen2022} proposed a localized operator inference approach in which multiple local reduced models are trained offline and an online classifier adaptively selects the most appropriate local ROM during prediction. 
In this model, the main adaptation mechanism is switching between pre-trained local ROMs rather than updating the operators online. 
More recently, Aretz and Willcox~\cite{aretz2025} introduced the nested operator inference method, which exploits hierarchical structure in the reduced space to enable efficient offline updates and warm-started refinements of learned operators. 
While these works demonstrate promising directions toward more flexible non-intrusive ROMs, neither performs continual online adaptation of model parameters.
Therefore, the present work introduces---to the best of our knowledge---the first set of ideas towards a general framework for adaptive non-intrusive reduced-order modeling. 

Our framework combines principles of online data assimilation and continual learning with reduced-order modeling to enable models that can refine themselves on-the-fly. 
Building upon the OpInf and NiTROM foundations, we develop three concrete adaptive formulations: (i)~Adaptive OpInf, which performs sequential updates to the basis and operators; (ii)~Adaptive NiTROM, which executes joint manifold optimization; and (iii)~Adaptive (hybrid) OpInf--NiTROM, which combines fast regression-based updates with short optimization-based refinements.
\reviewerone{It is important to note that all the proposed models assume periodic access to high-fidelity state snapshots online through single-step FOM solves. This deployment assumption affects the scope of applicability of the proposed framework.}
Through comprehensive numerical experiments on a canonical nonlinear flow problem, we assess the accuracy and practicality of these adaptive formulations. 
Our observations indicate that, when the system remains close to its training manifold, all tested adaptive models prevent amplitude drift and stabilize long-term predictions.
Moreover, when the system undergoes regime changes or is minimally trained offline, hybrid adaptation provides the most robust recovery and physically meaningful results.
These findings suggest that non-intrusive online adaptation has the potential to extend the predictive range of ROMs, marking a promising step toward \emph{truly} predictive reduced models. That said, we emphasize that the present frameworks are still at an early stage of development, and issues of computational efficiency remain open. It is the authors' opinion that the ROM community (including our own work) require  a stricter, cost-aware standard for claims of {\em prediction,} with clear separation of training and deployment regimes and full reporting of offline/online computational budgets. Reproducing previously seen dynamics may be useful in some circumstances, but it should not be presented as evidence of generalization absent validated out-of-manifold performance under stated resource and time constraints. We hope to provide some quantification in this sense.

The remainder of this paper is organized as follows. Section~\ref{sec:methods} presents problem formulation and the design choices in an adaptive non-intrusive ROM framework in detail. Section~\ref{sec:experimental_setup} outlines the full-order model, numerical setup, and evaluation metrics used to test the adaptive ROMs. Section~\ref{sec:results} reports and analyzes the results across progressively more challenging scenarios. Finally, Sec.~\ref{sec:conclusion} summarizes the main findings, discusses limitations, and outlines future directions.

%%%%%%%%%%%%%%%%%%%%%%%%%%%%%%%%%%%%%%%%%%%%%%%%%%%%%%%%%%%%%
\section{Methodology}
\label{sec:methods}
%%%%%%%%%%%%%%%%%%%%%%%%%%%%%%%%%%%%%%%%%%%%%%%%%%%%%%%%%%%%%

Consider a nonlinear high-dimensional dynamical system
\begin{equation}
  \begin{aligned}
      \frac{d}{dt}\mathbf{x}(t) = \mathbf{f}(\mathbf{x}(t),\mathbf{u}(t)) \ \ ; \ \ 
      \vecy(t) = \mathbf{h}(\vecx(t)),
  \end{aligned}
  \label{eq:FOM_general}
\end{equation}
with state $\vecx(t)\in\mathbb{R}^n$, control input $\vecu(t)\in\mathbb{R}^m$ and measured output $\vecy(t)\in\mathbb{R}^q$.
This system represents
the full-order model (FOM) that is often too expensive to simulate when the state dimension~$n$
is very large. 
A reduced-order model (ROM) is another dynamical system, usually \reviewerone{described} by the pipeline in Fig.~\ref{fig:rom_schematic}.
The key components of a ROM are~(i) an encoder, which is a possibly nonlinear map from the $n$-dimensional space to the $r$-dimensional space (with $r\ll n$),~(ii) a decoder, which is a possibly nonlinear map from the~$r$-dimensional space to the $n$-dimensional space, and~(iii) the latent-space dynamics $\vecf_r$ used to evolve the reduced-order state $\vecz$.
In this paper, we constrain our attention to the case where the encoder and decoder are linear maps, and the encoder is a left inverse of the decoder. 
Then, we may express the encoder as $\psi = \Psi^\top$, and the decoder as $\varphi = \Phi (\Psi^\top\Phi)^{-1}$, where $\Phi,\,\Psi\in\mathbb{R}^{n\times r}$ define, respectively, the
decoder basis (spanning the \emph{trial} space) and the encoder basis
(spanning the \emph{test} space).
By virtue of the fact that the encoder is a left inverse of the decoder, the encoder/decoder pair define a projection operator
\begin{equation}
\label{eq:proj_P}
  \mathbb{P} \triangleq \Phi (\Psi^\top\Phi)^{-1}\Psi^\top:\mathbb{R}^n\to\mathbb{R}^n,\quad \mathbb{P}^2 = \mathbb{P}.
\end{equation}
The ROM in Fig.~\ref{fig:rom_schematic} approximates the FOM solution in the range of $\mathbb{P}$.
When $\text{span}(\Phi) \equiv \text{span}(\Psi)$, we say that the projection is orthogonal, and it is oblique otherwise.
We remark that encoders and decoders can also be defined by nonlinear functions~\citep{lee2020, fresca2021, conti2023, otto2023ae}, although strongly enforcing that the encoder $\psi$ be a left inverse of the decoder $\varphi$ requires some care.
In any case, this is beyond the scope of this work.

\begin{figure}
\centering
\scalebox{0.9}{
  \begin{tikzpicture}
    [
    trap/.style ={trapezium,draw,minimum height=1cm,minimum width=2cm}
    ]
    % Draw control volume 
    \node [trapezium,draw,rotate=-90,minimum width=2cm,minimum height=1cm,rounded corners=1mm,label=center:\textcolor{black}{$\psi$}](psiTrap) at (0,0){};
    \node [trapezium,draw,rotate=90,minimum width=2cm, minimum height=1cm,rounded corners=1mm,right=6.9 of psiTrap,label={[rotate=0]center:\textcolor{black}{$\varphi$}}](phiTrap){};
    \node [left=1.8 of psiTrap.center,label={[xshift=0.5cm, yshift=0.0cm]$\vecx(0)\in \mathbb{R}^n$}] (1) {};
    \draw [->,-stealth] (1) -- (psiTrap.270);
    \node [right=1.7 of phiTrap.center,label={[xshift=-0.45cm, yshift=0.0cm]$\hat{\vecx}(t)\in \mathbb{R}^n$}] (2) {};
    \draw [->,-stealth] (phiTrap.270) -- (2);
    \node [rectangle,draw,right=2.25 of psiTrap.center,minimum height=1cm,rounded corners=1mm](rect) {\textcolor{black}{$\frac{d}{dt}\vecz = \vecf_r(\vecz,\vecu)$}};
    \draw [->,-stealth] (psiTrap.90) -- (rect.180);
    \draw [->,-stealth] (rect.0) -- (phiTrap.90);
    \node [below=0.5 of rect.270](u){$\vecu$};
    \draw [->,-stealth] (u) -- (rect.270);
    \node [right=0.7 of psiTrap.90,label={[xshift=0.05cm, yshift=0.0cm]$\vecz(0)\in\mathbb{R}^r$}] {};
    \node [right=0.62 of rect,label={[xshift=0.1cm, yshift=0.0cm]$\vecz(t)\in\mathbb{R}^r$}] {};
  \end{tikzpicture}}
\caption{Schematic of a ROM with encoder \textcolor{black}{$\psi$}, decoder \textcolor{black}{$\varphi$} and latent-space dynamics \textcolor{black}{$\vecf_r(\vecz,\vecu)$}.}
\label{fig:rom_schematic}
\end{figure}

The latent-space dynamics $\vecf_r$ in Fig.~\ref{fig:rom_schematic} can be defined in several ways.
Perhaps the most well-known is via (Petrov-)Galerkin projection, where given a projection operator $\mathbb{P}$ defined as in (\ref{eq:proj_P}), the latent-space dynamics is obtained by projecting the full-order dynamics $\vecf$ as follows
\begin{equation}
    \vecf_r(\vecz, \vecu) = \Psi^\top\mathbf{f}(\Phi(\Psi^\top\Phi)^{-1}\mathbf{z},\mathbf{u}).
\end{equation}
This procedure is generally considered (weakly) intrusive, because it requires direct access to the full-order dynamics $\vecf$.
Alternatively, the latent-space dynamics can be inferred from data using non-intrusive approaches that require no access to $\vecf$.
For instance, this is the case of OpInf (discussed in greater detail in the following section), where, assuming dynamics of polynomial form
\begin{equation}
  \mathbf{f}_r(\mathbf{z},\mathbf{u})
  = A_r\mathbf{z}
    + H_r:(\mathbf{z}\mathbf{z}^{\top})
    + B_r\mathbf{u} + \ldots,
  \label{eq:poly_reduced}
\end{equation}
the reduced-order tensors $A_r\in\mathbb{R}^{r\times r}$, $H_r\in\mathbb{R}^{r\times r\times r}$, $B_r\in\mathbb{R}^{r\times m}$, etc. are learned from data by solving a least-squares optimization problem.

Clearly, whether we are operating in a static setting or in an adaptive setting, the encoder, decoder, and latent-space dynamics  determine the accuracy of the ROM and its ability to forecast the system's dynamics.
In this paper, we consider adaptive, \emph{non-intrusive} models with polynomial latent-space dynamics as in Eq.~\eqref{eq:poly_reduced}, and with linear encoders/decoders defining (possibly oblique) projectors as in Eq.~\eqref{eq:proj_P}.
In particular, we will consider three adaptive paradigms: one based on OpInf, one on NiTROM, and one on a combination of these.
Before presenting the adaptive methodologies, it is instructive to review their static counterparts.

%%%%%%%%%%%%%%%%%%%%%%%%%%%%%%%%%%%%%%%%%%%%%%%%%%%%%%%%%%%%%
\subsection{Static ROMs}
\label{subsec:static_models}
%%%%%%%%%%%%%%%%%%%%%%%%%%%%%%%%%%%%%%%%%%%%%%%%%%%%%%%%%%%%%

\subsubsection{Operator Inference}
\label{subsec:static_opinf}

As previously mentioned, Operator Inference (OpInf)~\citep{peherstorfer2016opinf,kramer2024opinf} is a non-intrusive model reduction framework where one usually seeks a polynomial ROM evolving on the range of an \emph{orthogonal} projection defined by (orthonormal) Proper Orthogonal Decomposition (POD) modes, $\Phi$.
In the terminology of Fig.~\ref{fig:rom_schematic}, the encoder is defined as $\psi = \Phi^\top$, the decoder as $\varphi = \Phi$, and the latent-space dynamics $\vecf_r$ as in Eq.~\eqref{eq:poly_reduced}.
Mathematically, given the leading $r$ orthonormal POD modes $\Phi\in\mathbb{R}^{n\times r}$ computed from a high-fidelity training data set $\{\vecx^{(i)}(t_j)\}$ (where $\vecx^{(i)}(t_j)\in\mathbb{R}^n$ is the solution vector at time $t_j$ along trajectory $i$), we compute the latent-space coordinates $\vecz = \Phi^\top \vecx$ and their time derivatives $d\vecz/dt = \Phi^\top d\vecx/dt$, and we solve the following least-squares optimization problem
\begin{equation}
    \min_{\Theta\in\mathcal{M}_{\text{OpInf}}}J = \frac{1}{ML}\sum_{i=0}^{L-1}\sum_{j=0}^{M-1} \bigg\lVert \frac{d}{dt}\vecz^{(i)}(t_j) - \vecf_r(\vecz^{(i)}(t_j),\vecu^{(i)}(t_j))  \bigg\rVert^2,\quad \Theta = (A_r, H_r, B_r,\ldots),
    \label{eq:opinf_cost}
\end{equation}
where $\vecf_r$ is defined as in Eq.~\eqref{eq:poly_reduced} and $\mathcal{M}_{\text{OpInf}} = \mathbb{R}^{r\times r} \times \mathbb{R}^{r\times r\times r} \times \mathbb{R}^{r\times m}\times \ldots$.
(Here, $L$ and $M$ are the number of training trajectories and training snapshots per trajectory, respectively.)
While OpInf is an elegant formulation with a closed-form solution, it was recently shown~\citep{padovan2024} that it often struggles to accurately forecast trajectories through large-amplitude transients.
This is because, although optimal at statically reconstructing the state, the underlying orthogonal projection $\mathbb{P}=\Phi\Phi^\top$ onto POD modes can accidentally truncate low-energy features that are nonetheless important for the dynamical evolution of the system.

\subsubsection{NiTROM}
\label{subsec:static_nitrom}

The  NiTROM framework recently introduced in Padovan et al.~\citep{padovan2024} is a non-intrusive formulation developed to address the shortcomings of non-intrusive models obtained by orthogonal projection onto high-energy/high-variance subspaces.
More specifically, leveraging the fact that \emph{oblique} projections have been shown to be better suited for capturing the dynamically-relevant mechanisms of FOMs exhibiting large-amplitude transient growth, NiTROM \emph{simultaneously} seeks optimal encoders $\psi = \Psi^\top$, decoders $\varphi = \Phi\left(\Psi^\top\Phi\right)^{-1}$, and polynomial latent-space dynamics~\eqref{eq:poly_reduced} by solving an optimization problem on a matrix manifold.
Given training data $\{\vecx^{(i)}(t_j)\}$, where, once again, $\vecx^{(i)}(t_j)\in\mathbb{R}^n$ is the solution vector at time $t_j$ along trajectory $i$, we solve the following problem
\begin{equation}
\label{eq:nitrom_opt_problem}
    \begin{aligned}
        \min_{\Theta \in \mathcal{M}_{\text{NiTROM}}}\,\,\, J &=  \frac{1}{ML}\sum_{i=0}^{L-1}\sum_{j=0}^{M-1}\lVert \vecy^{(i)}(t_j) - \hat{\vecy}^{(i)}(t_j)\rVert^2 \\
        \text{subject to: } \quad \frac{d}{d t}\vecz^{(i)} &=  \vecf_r(\vecz^{(i)},\vecu^{(i)}),\quad \vecz^{(i)}(t_0) = \Psi^\intercal \vecx^{(i)}(t_0) \\
            \hat{\vecy}^{(i)} &= \mathbf{h}\left(\Phi\left(\Psi^\top \Phi\right)^{-1} \vecz^{(i)}\right)\\
            V &= \text{Range}\left(\Phi\right),
    \end{aligned}
\end{equation}
\reviewerthree{with $\vecy^{(i)}(t_j)=\mathbf{h}\!\left(\vecx^{(i)}(t_j)\right)$}, parameters $\Theta = \left(V,\Psi,A_r, H_r, B_r,\ldots\right)$, and $\mathcal{M}_{\text{NiTROM}}$ a matrix manifold defined as 
\begin{equation}
    \mathcal{M}_{\text{NiTROM}} = \mathcal{G}_{n,r}\times S_{n,r} \times \mathbb{R}^{r\times r} \times \mathbb{R}^{r\times r\times r} \times \mathbb{R}^{r\times m}\times \ldots.
\end{equation}
Here, $\mathcal{G}_{n,r}$ is the Grassmann manifold (i.e., the space of $r$-dimensional subspaces of $\mathbb{R}^n$) and $S_{n,r}$ is the Stiefel manifold (i.e., the space of $n\times r$ orthonormal matrices).
This particular parameterization arises from the nature of the optimization problem.
In particular, we optimize subspaces over the Grassmann manifold because the optimization problem~\eqref{eq:nitrom_opt_problem} is a function of the subspace $V$ spanned by the decoder $\Phi\left(\Psi^\top\Phi\right)^{-1}$, rather than of the matrix representative $\Phi\left(\Psi^\top\Phi\right)^{-1}$ itself.
The encoder $\Psi^\top$, on the other hand, is optimized over the Stiefel manifold to strongly enforce the necessary condition $\text{rank}(\Psi) = r$ in order to define a projection $\mathbb{P}$ as in Eq.~\eqref{eq:proj_P}.
The other (linear) spaces (e.g., $\mathbb{R}^{r\times r}$, etc.) in the definition of $\mathcal{M}_{\text{NiTROM}}$ are used to parameterize the reduced-order tensors defining the latent-space polynomial dynamics in Eq.~\eqref{eq:poly_reduced}.

This formulation has several advantages over the previously-discussed OpInf. 
First, by allowing for the encoder and decoder to span different spaces, the model is more expressive and can therefore detect the low-energy sensitivity mechanisms that are important for accurate forecasting.
Second, the cost function measures the solution error in the original high-dimensional space, rather than measuring deviations in the time-rate-of-change of the latent-space vector, ultimately encouraging stability of the latent-space dynamics.
Third, all ROM components (encoder, decoder, and dynamics) are optimized simultaneously, thus promoting the computation of \emph{dynamics-aware} projection operators leading to models that are capable of predicting the FOM solution along transient trajectories.
A drawback of this formulation, however, is that the optimization problem is non-convex, \reviewerthree{implying the existence of local minima and non-unique minimizers}.
Additionally, unlike OpInf, the problem does not admit a closed-form solution.
Nonetheless, it was shown in \citep{padovan2024} that the gradient of the cost function $J$ with respect to the parameters $\Theta$ can be computed in closed-form using a fully non-intrusive adjoint-based formulation.
This implies that the optimization problem can be solved efficiently without relying on possibly expensive automatic differentiation.

%%%%%%%%%%%%%%%%%%%%%%%%%%%%%%%%%%%%%%%%%%%%%%%%%%%%%%%%%%%%%
\subsection{Adaptive ROMs}
\label{subsec:adaptive_philosophy}
%%%%%%%%%%%%%%%%%%%%%%%%%%%%%%%%%%%%%%%%%%%%%%%%%%%%%%%%%%%%%

\begin{figure}[t]
  \centering
  \includegraphics[width=0.75\linewidth]{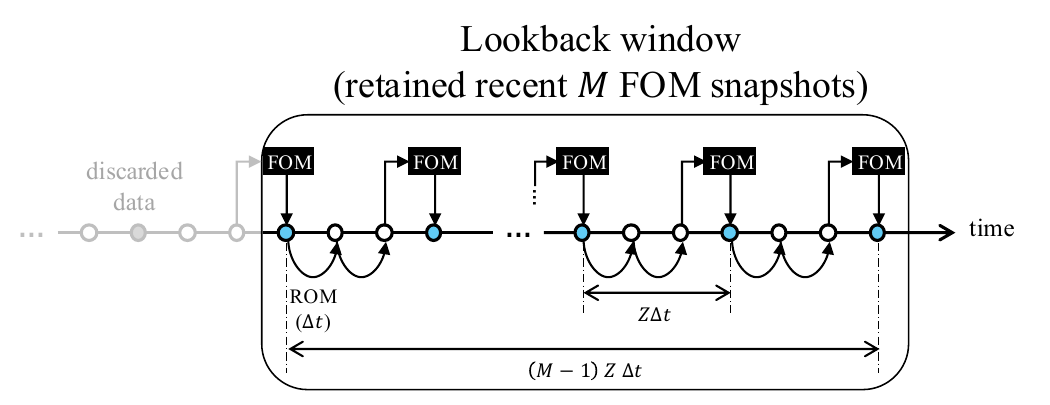}
  \caption{Schematic for the adaptation setup. Highlighted steps show FOM snapshots kept in the moving window.}
  \label{fig:adaptation_schematic}
\end{figure}

Adaptive model reduction methods rely on  interactions with the full-order, high-fidelity solver to update the model parameters and maintain a desired level of accuracy.
The adaptation setup used throughout this paper is best described graphically in Fig.~\ref{fig:adaptation_schematic}.
Here, $\Delta t$ is the FOM time step, and $Z\Delta t$ is the length of the ROM forecast window.
In other words, time advancement from time $t_j - Z\Delta t$ to $t_j$ is performed using the reduced-order model.
Every $Z\Delta t$ time units, the ROM interacts with the FOM and collects a new sample of high-fidelity data, which is stored in a data matrix
\begin{equation}
  \mathcal{D}_j
  =
  \big\{
    Q(t_{j-(M-1)Z}),\;
    \ldots,\;
    Q(t_{j-Z}),\; Q(t_j)
  \big\},\quad Q(t_i) = \left(\mathbf{x}(t_i),\dot{\mathbf{x}}(t_i),\mathbf{u}(t_i)\right).
  \label{eq:window}
\end{equation}
 $\mathcal{D}_j$ is a moving-window data matrix since, at any given time $t_j$, it contains~$M$ high-fidelity samples, and when a new sample becomes available, the ``oldest'' sample is discarded.
This way, older samples that are no longer representative of the current state of the system are naturally forgotten, enabling the model to emphasize the most
recent and relevant patterns. \reviewerthree{It is worth mentioning that performing multiple consecutive FOM steps per adaptation
is also possible, and that the single-step choice here reflects a cost-aware design decision.}

The goal of each adaptation is to update the ROM parameters
$\Theta_{j-1}$ to a refined set~$\Theta_j$ that better represents the
newly observed dynamics contained in $\mathcal{D}_j$.
This update is the solution of a
localized learning problem of the form
\begin{equation}
  \Theta_j
  =
  \arg\min_{\Theta\in\mathcal{M}}
  J_j(\Theta),
  \label{eq:window_objective}
\end{equation}
where $J_j$ is some objective cost function formulated over the training data in $\mathcal{D}_j$.
For instance, when the model is adapted according to an OpInf-based scheme, the cost function $J_j$ is given by the cost function $J$ in Eq.~\eqref{eq:opinf_cost} with $L = 1$, and $\mathcal{M} = \mathcal{M}_{\text{OpInf}}$.
Similarly, when the model is adapted using a NiTROM-like rule, the cost function is given by Eq.~\eqref{eq:nitrom_opt_problem}, also with $L=1$ and with $\mathcal{M}=\mathcal{M}_{\text{NiTROM}}$.

Adaptations are triggered at discrete intervals during online
integration.
In the simplest case, updates occur every $Z$ time steps.
This fixed schedule simplifies
analysis and allows direct control over computational cost.
An alternative strategy could employ adaptive triggering based on
residual norms, which is a natural
extension for future work.
The three main hyperparameters controlling the overall behavior of the
adaptive ROM are therefore
\(
  (M,Z,K),
\)
where $M$ is the number of high-quality snapshots retained in the moving (lookback)
window, $Z$ is the number of reduced time steps between successive
adaptations (adaptation window), and $K$ is the
number of optimization steps (if doing gradient-based online learning, as in the NiTROM framework)
per adaptation.
These quantities jointly control the trade-off between accuracy
and computational efficiency.
The adaptation framework is described in pseudocode in Alg.~\ref{alg:unified_framework}, where we describe the online forecasting phase and the ROM-FOM interaction.

\begin{remark}
    \reviewerone{We emphasize that decoding a ROM-predicted state produces a state in the decoder range, which, in general, does not span the FOM solution manifold. 
    The role of periodic FOM interaction is to correct accumulated model error before it grows irreversibly, by injecting newly observed dynamics back into the reduced model. In practice, this requires that the ROM prediction remains sufficiently close to the true trajectory between adaptation events, a condition that is controlled by the adaptation window $Z$. Our numerical results in Sec.~\ref{sec:results} demonstrate that when this condition is respected, a one-step FOM advancement from the lifted state produces physically consistent snapshots that enable effective adaptation.}
\end{remark}

\begin{algorithm}[t]
\caption{Adaptive Non-Intrusive ROM Framework}
\label{alg:unified_framework}
\begin{algorithmic}[1]
\Require Initial training data set $\mathcal{D}_{0}$ with $M$ samples sampled at times $t\in [t_0 - (M-1)Z\Delta t, t_0]$; reduced dimension $r$; adaptation window $Z$; lookback window $M$; optimization budget $K$ (if applicable);
\State $j \gets 0$
\State $\Theta_j \gets \textsc{InitializeModel}(\mathcal{D}_{j}, r)$ 

\For{$t \in t_0 + \Delta t\{0, 1, 2,\ldots\}$}
  \State $\vecz(t)\gets \textsc{AdvanceROMForOneTimeStep}(\Theta_j, \vecz(t-\Delta t))$
  \If{$(t - t_0 +\Delta t) \bmod (Z \Delta t) = 0$}
  \State Decode $\hat{\vecx}(t - \Delta t)\gets \Phi_j\left(\Psi_j^\top\Phi_j\right)^{-1}\vecz(t - \Delta t)$
  \State $\vecx(t)\gets \textsc{AdvanceFOMForOneTimeStep}(\hat{\vecx}(t-\Delta t))$
  \State $\mathcal{D}_{j+1}\gets \textsc{UpdateDataMatrix}(\mathcal{D}_{j},\vecx(t))$
      \State $\Theta_{j+1} \gets \textsc{AdaptModel}(\Theta_j, \mathcal{D}_{j+1}, Z, M, K)$
      \State $j \gets j + 1$
  \EndIf
\EndFor
\end{algorithmic}
\end{algorithm}

%%%%%%%%%%%%%%%%%%%%%%%%%%%%%%%%%%%%%%%%%%%%%%%%%%%%%%%%%%%%%
\subsubsection{Adaptive Operator Inference}
\label{subsec:adaptive_opinf}
%%%%%%%%%%%%%%%%%%%%%%%%%%%%%%%%%%%%%%%%%%%%%%%%%%%%%%%%%%%%%

Adaptive OpInf decomposes the online update in Eq.~\eqref{eq:window_objective} into two sequential stages:  
(1) update the reduced basis using the new information contained in the
moving window~$\mathcal{D}_j$, and  
(2) refit the reduced operators by solving a least-squares problem on
the same window.

\paragraph{Stage~1: Basis update.}
At the $j$th adaptation event, the most recent $M$ high-quality
snapshots form the data window
$\mathcal{D}_j$ as defined in Eq.~\eqref{eq:window}.
As $\mathcal{D}_j$ is updated in time, its span changes, and so should the POD modes $\Phi$ used for OpInf (see Sec.~\ref{subsec:static_opinf} for details).
Several strategies are available to update the basis.
\begin{itemize}
  \item \emph{Windowed SVD:}  
        Compute the singular value decomposition (SVD) over the window
        and select the $r$ leading left singular vectors. This approach
         ignores the previous basis and re-builds the basis from
        scratch at every adaptation.
        This provides the optimal rank-$r$ subspace (in the
        $\ell_2$ sense) for the current window, but requires a full SVD
        at every adaptation. This method has been used throughout our experiments
        to establish a solid baseline, even though more efficient
        techniques (described below) exist.
  \item \emph{Incremental SVD (iSVD):}  
        Update the basis with the new snapshot $\mathbf{x}(t_j)$, 
        using recursive low-rank updates. This provides an efficient, incremental
        technique that avoids recomputing the SVD (\reviewerthree{See, e.g., Brand's
        iSVD algorithm~\citep{brand2002}}).
  \item \emph{Rank-1 update:}  
        The approach of Huang \& Duraisamy~(2023)~\citep{huang2023} provides
        an efficient rank-one correction to the old basis that effectively removes the projection
        error from the new snapshot. This is a rather aggressive approach as old snapshots have no direct contributions to the new basis.
        \reviewerthree{The method is effective when spatially localized updates are made to the basis at each time step, which
        typically requires intrusive access to the FOM operators at selected points in the domain. This operation is not feasible in the non-intrusive setting considered here, and therefore, rank-one updates are not explored in this work.}
\end{itemize}

\paragraph{Stage~2: Operator refit.}
Once the basis $\Phi_j$ has been updated, the reduced operators
$\Theta_j = ({A_r}_j,{H_r}_j,{B_r}_j)$ are re-computed by solving the optimization problem outlined in Sec.~\ref{subsec:static_opinf} with $L = 1$.
The sequential two-stage update of Adaptive OpInf provides a
lightweight yet effective mechanism for online learning
of model parameters.

\begin{remark}
\reviewerthree{
In the OpInf setting, uniqueness of the least-squares solution requires
the regression data matrix to have full column rank, which typically imposes a lower bound
on the number of snapshots relative to the number of operator coefficients. In the present
adaptive setting, we could face cases where the number of snapshots kept in the moving window does not satisfy the full column rank condition. To improve conditioning in these cases, we employ regularized least squares,
which yields a well-posed solution suitable for short-horizon prediction.
}
\end{remark}

%%%%%%%%%%%%%%%%%%%%%%%%%%%%%%%%%%%%%%%%%%%%%%%%%%%%%%%%%%%%%
\subsubsection{Adaptive NiTROM}
\label{subsec:adaptive_nitrom}
%%%%%%%%%%%%%%%%%%%%%%%%%%%%%%%%%%%%%%%%%%%%%%%%%%%%%%%%%%%%%

Adaptive NiTROM extends the offline NiTROM optimization framework to the online,
streaming context.
Unlike Adaptive OpInf, which treats the basis and reduced operators as
separate entities, Adaptive NiTROM performs a
joint optimization, as discussed in Sec.~\ref{subsec:static_nitrom}.
This enables coherent corrections of both the reduced subspaces and the
latent dynamics in a single optimization step, ensuring that the basis
and operators remain mutually consistent with the evolving data.
Formally, at each adaptation event, we solve the NiTROM problem in Sec.~\ref{subsec:static_nitrom} against the training data in $\mathcal{D}_j$.
The optimization is warm-started from the previous parameters
$\Theta_{j-1}$.
The adaptive update proceeds using Riemannian gradient-based steps on
the product manifold $\mathcal{M}_{\text{NiTROM}}$.

%%%%%%%%%%%%%%%%%%%%%%%%%%%%%%%%%%%%%%%%%%%%%%%%%%%%%%%%%%%%%
\subsubsection{Adaptive (Hybrid) OpInf--NiTROM}
\label{subsec:hybrid}
%%%%%%%%%%%%%%%%%%%%%%%%%%%%%%%%%%%%%%%%%%%%%%%%%%%%%%%%%%%%%

While Adaptive NiTROM provides the most principled framework for
trajectory-level adaptation, its convergence and cost depend strongly on
the initialization point~$\Theta_{j-1}$.
If the system dynamics have changed significantly since the last
adaptation, starting directly from~$\Theta_{j-1}$ can lead to poor
local minima or slow convergence.
On the other hand, Adaptive OpInf can rapidly compute an approximate
update using least squares on the new data window, but lacks the
geometry-aware coupling between basis and operators.
The central idea of a hybrid approach, which we call Adaptive OpInf--NiTROM,
is to use a fast OpInf update as a
\emph{jump-start} for the subsequent manifold optimization, thereby providing a high-quality initialization that places the
optimization algorithm near a favorable region of the cost landscape.

%%%%%%%%%%%%%%%%%%%%%%%%%%%%%%%%%%%%%%%%%%%%%%%%%%%%%%%%%%%%%
\subsection{Computational considerations}
\label{subsec:computation}
%%%%%%%%%%%%%%%%%%%%%%%%%%%%%%%%%%%%%%%%%%%%%%%%%%%%%%%%%%%%%

The computational cost of an adaptive ROM is a critical factor in
assessing its practicality.
While model reduction is motivated by the desire for faster prediction,
adaptivity introduces additional overhead during the online phase.
Here we outline the computational scaling of each major component of
the proposed adaptive framework.

Between two successive adaptations, the ROM is integrated entirely in
the reduced space of dimension~$r$.
For a ROM with a polynomial nonlinearity of degree $\alpha$,
the number of distinct degree-$\alpha$ monomials in $r$ variables is
$s=\binom{r+\alpha-1}{\alpha}=\mathcal{O}(r^{\alpha})$, and therefore
the evaluation of the right-hand side of Eq.~\eqref{eq:poly_reduced} at each time step scales as
$\mathcal{O}(r^{\alpha+1})$,
dominated by the highest-degree term.
Since $r \!\ll\! n$, this cost is negligible compared with both the
high-fidelity solver and the subsequent adaptation step.

At each adaptation event, a new high-quality snapshot must be obtained.
In a digital twin setting, this corresponds to sensor data and therefore
incurs no computational cost.
In simulation-based environments, however, a new snapshot is obtained by
lifting the ROM prediction to the full space and performing a one-step
FOM advancement.
The lifting cost scales as $\mathcal{O}(nr)$, while the FOM advancement
typically scales as $\mathcal{O}(n^{\beta})$, where $\beta > 1$ for implicit schemes,
depending on the solver type and sparsity structure of the system matrices.
In realistic CFD solvers, this one-step FOM query often dominates the
total cost of an adaptation cycle.

After acquiring the new snapshot, the ROM parameters are updated using
one of the adaptive formulations discussed previously.
For Adaptive OpInf, the total per-update cost includes the basis update,
the projection between full and reduced spaces, and the least-squares
refit of the reduced operators.
The basis update can be performed using several strategies: a windowed
SVD with cost $\mathcal{O}(n M^2)$, an incremental SVD (iSVD) with
$\mathcal{O}(n r)$ cost, or a rank-one update with similar scaling to iSVD.
Projecting the data to and from the reduced space costs
$\mathcal{O}(M n r)$ per window.
For a polynomial model of degree $\alpha$,
assembling the OpInf data matrices over $M$ snapshots incurs
$\mathcal{O}(Mr^{\alpha})$ work. The least-squares refit then has
an approximate upper bound of
$
\mathcal{O}(Mr^{3\alpha+1}),
$
which reflects the exponential growth in $\alpha$ (see Sec.~3.5 of
\citep{peherstorfer2016opinf}).
Putting these pieces together, a single Adaptive OpInf update (with windowed SVD) scales as
\[
\mathcal{O}(\reviewerone{nM^2} + M n r + Mr^{3\alpha+1}).
\]

The Adaptive NiTROM model, by contrast, involves a joint manifold
optimization with $K$ Riemannian iterations per adaptation.
Each iteration requires one forward and one adjoint integration of the
reduced model across a window of $M$ time steps.
The corresponding cost scales as
$\mathcal{O}(K M r^{\alpha+1})$, where $\alpha$ is the polynomial degree of the
ROM.
Additional terms arise from \reviewertwo{matrix-vector products} involving the encoder and decoder
bases $(\Phi,\Psi)$ on the Stiefel and Grassmann manifolds, which approximately add
$\mathcal{O}(K n r^2)$ per update (see Sec.~2.4 of \citep{padovan2024}).
Hence, the total per-adaptation complexity of Adaptive NiTROM is
\[
\mathcal{O}(K n r^2 + K M r^{\alpha+1}).
\]

The Adaptive OpInf--NiTROM method combines the two mechanisms by first
performing a rapid OpInf update, followed by a small number of
NiTROM refinement steps.
Its total cost can be estimated as
\[
  \mathcal{O}(\reviewerone{nM^2} + M n r + Mr^{3\alpha+1})
  + \mathcal{O}(K n r^2 + K M r^{\alpha+1}).
\]
In practice, the cost is dominated by the number of NiTROM iterations,
which is typically small ($K\sim5$--$10$) since the OpInf stage
provides a near-optimal initialization.

Table~\ref{tab:cost_summary} summarizes the asymptotic per-adaptation
costs of all adaptive models considered in this work. Note that only the terms scaling with the full dimension $n$ are kept, as all other terms are negligible.
A quantitative comparison is reported in Sec.~\ref{sec:results} (see Table~\ref{tab:wallclock}), where wall-clock time of each step is presented.
In general, the computational overhead introduced by adaptation remains
small relative to high-fidelity simulation, but it is non-negligible for frequent updates or several optimization steps, and thus represents a practical
constraint on real-time deployment.

\begin{table}[t]
  \centering
  \caption{Asymptotic computational cost per adaptation. Here, $n$ is the full dimension, $r$ the reduced dimension, $\alpha$ the polynomial degree of the reduced operators, $M$ the number of snapshots in the moving window, and $K$ the number of optimization iterations.}
  \label{tab:cost_summary}
  \begin{tabular}{lcc}
    \toprule
    \textbf{Model} & \textbf{Operation} & \textbf{Cost} \\
    \midrule
    % Static Galerkin / OpInf / NiTROM & ROM time integration & $\mathcal{O}(r^3)$ per step \\
    Adaptive OpInf & Basis + operator update & $\mathcal{O}\big(M n r)$ \\
    % Adaptive OpInf (iSVD / rank-1) & Basis + operator update & $\mathcal{O}(n r + m r^3)$ \\
    Adaptive NiTROM & Joint manifold update & $\mathcal{O}(Kn r^2)$ \\
    Adaptive OpInf--NiTROM & Hybrid & $\mathcal{O}(n(\reviewerone{M^2} + Mr + Kr^2))$ \\
    \midrule
    FOM one-step advancement & High-fidelity solve & $\mathcal{O}(n^{\beta}), \ \beta > 1$ \\
    \bottomrule
  \end{tabular}
\end{table}

%%%%%%%%%%%%%%%%%%%%%%%%%%%%%%%%%%%%%%%%%%%%%%%%%%%%%%%%%%%%%
\section{Experimental setup}
\label{sec:experimental_setup}
%%%%%%%%%%%%%%%%%%%%%%%%%%%%%%%%%%%%%%%%%%%%%%%%%%%%%%%%%%%%%

\subsection{Full-order model (FOM)}
Our test problem is the canonical lid-driven cavity flow at Reynolds number $\mathrm{Re}=8300$,
identical to that in Sec.~5 of~\citep{padovan2024}.
The system is governed by the two-dimensional incompressible Navier--Stokes equations in non-dimensional form
\begin{align}
  \frac{\partial \mathbf{V}}{\partial t} + \mathbf{V}\cdot\nabla\mathbf{V}
  &= -\nabla P + \mathrm{Re}^{-1}\nabla^2\mathbf{V},
  \label{eq:ns_fom}\\
  \nabla\cdot\mathbf{V}&=0,
  \nonumber
\end{align}
where $\mathbf{V} = (U, V)$ is the two dimensional velocity vector, $P$ is the pressure, and the solution domain is the unit square $D=[0,1]\times[0,1]$ with zero velocity at all
walls except $U=1$ at the top lid.
Spatial discretization is performed using a second-order finite-volume
scheme on a $100\times100$ fully staggered uniform grid, through which the projection step
automatically enforces the divergence-free constraint and removes
pressure from the state variables.
The resulting semi-discrete system contains
$n=2\times10^4$ unknowns.
In our experiments, the observation operator $\mathbf{h}$ (see Eq.~\eqref{eq:FOM_general}) is chosen to be the identity,
giving the full velocity field $\mathbf{v}$ as the output.

\begin{figure}[t]
  \centering
  \includegraphics[trim=0pt 0pt 0pt 0pt, clip, width=\linewidth]{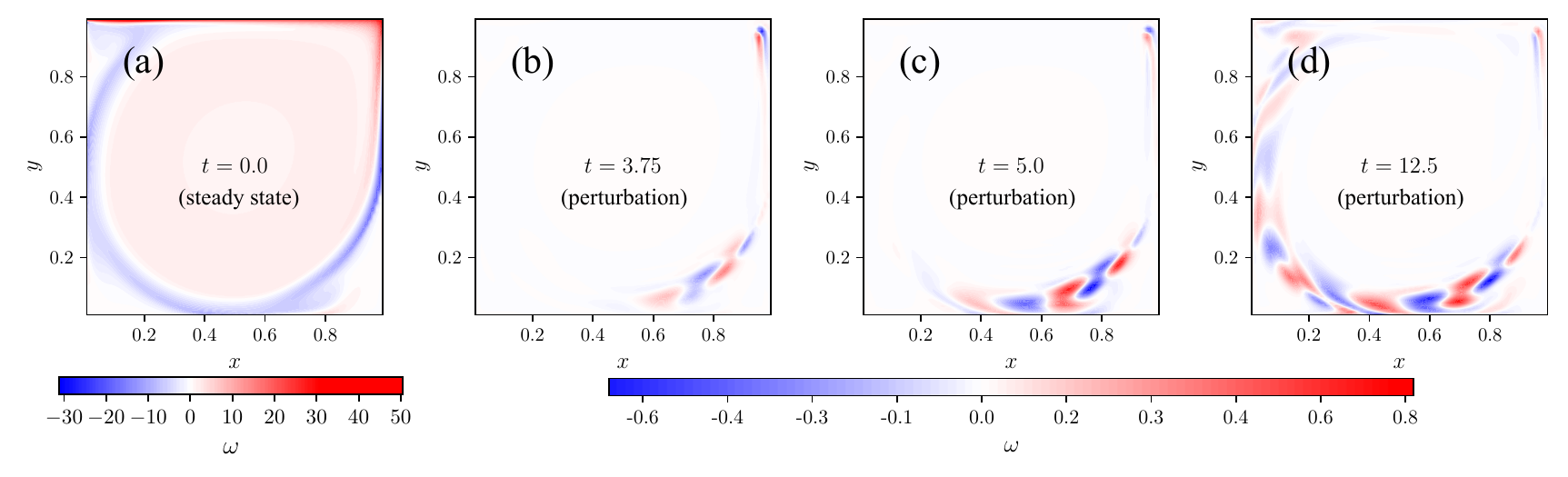}
  \caption{{Transient 2-D lid-driven cavity flow. The leftmost field depicts the vorticity field from the steady state solution of the system at $Re=8300$, and the remaining plots show vorticity fluctuations, showing how the dynamics evolve in time.}}
  \label{fig:FOM}
\end{figure}

The system is driven by a spatially-localized input acting on the
$x$-momentum equation,
\begin{equation}
  B(x,y)\,w(t)
  = \exp\!\big\{-5000\big((x-x_c)^2+(y-y_c)^2\big)\big\}\,w(t),
  \quad x_c=y_c=0.95,
  \label{eq:input_map}
\end{equation}
where $B(x,y)$ defines the input map and $w(t)$ is a scalar forcing function.
Time integration uses a two-step Adams-Bashforth scheme with step
$\Delta t = 0.0025$ to retain stability.
The flow admits a stable steady state $\overline{\mathbf{V}}$ at this Reynolds number but
exhibits strong transient growth when perturbed.
We initialize the simulation from the steady state and excite the system
with a sinusoidal forcing,
\begin{equation}
    \label{eq:u}
  w(t) = 0.1\,\sin(4t),
\end{equation}
which injects momentum near the upper-right corner of the cavity, at $x_c=y_c=0.95$, as mentioned in Eq.~\eqref{eq:input_map}.
This continuous forcing generates rich transient and
oscillatory behavior (Fig.~\ref{fig:FOM}), providing a suitable benchmark for testing adaptive
ROMs.

\begin{remark}
    This problem was chosen because it  exhibits a range of features (e.g., transient growth and traveling-wave-like structures) that appear in less canonical flow configuration and pose a challenge to the development of accurate ROMs.
    Ablation studies and extensive numerical experiments in different training/testing regimes expose strengths and weaknesses of the proposed non-intrusive adaptive ROMs approaches, and shine a light on the research questions that must be addressed before these ROMs can be successfully deployed on more large-scale problems.
\end{remark}

\subsection{Training and test windows}
\begin{figure}[tb!]
    \centering
    \begin{subfigure}{0.33\textwidth}
        \includegraphics[width=\textwidth]{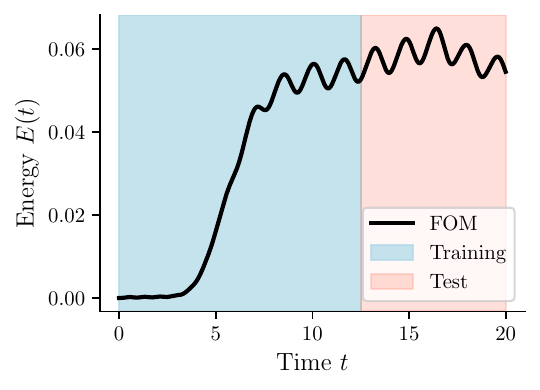}
        \vspace{-2em}
        \caption{{Case~1: rich training}}
        \label{subfig:case1}
    \end{subfigure}
    \hspace{-1.em}
    \begin{subfigure}{0.33\textwidth}
        \includegraphics[width=\textwidth]{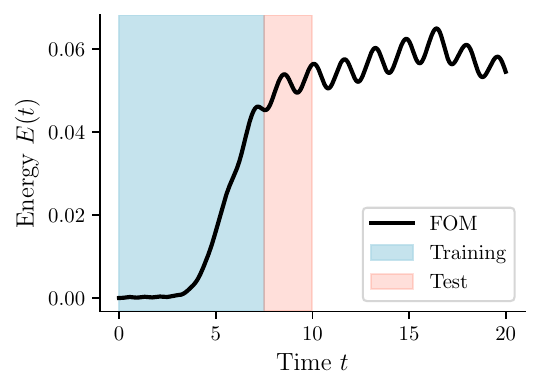}
        \vspace{-2em}
        \caption{{Case~2: regime change}}
        \label{subfig:case2}
    \end{subfigure}
    \hspace{-1.em}
    \begin{subfigure}{0.33\textwidth}
        \includegraphics[width=\textwidth]{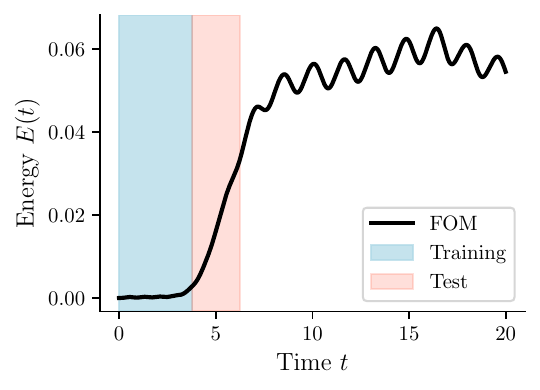}
        \vspace{-2em}
        \caption{{Case~3: minimal training}}
        \label{subfig:case3}
    \end{subfigure}
    
    \caption{Illustration of the three training/test window configurations. 
Each panel shows the temporal evolution of the system energy, 
\reviewerthree{with the blue-shaded region denoting the offline (training) window $[t_0,t_1]$ 
and the red-shaded portion showing the online (test) window $(t_1,t_2]$.}
The training interval progressively shortens from case~1 
to case~3.}
    \label{fig:cases}
\end{figure}

Throughout the rest of the paper, we test our adaptive frameworks in three different regimes with progressively shorter offline training windows.
When we force the flow from rest with the forcing signals in Eqs.~\eqref{eq:input_map} and~\eqref{eq:u}, the flow undergoes a sharp transient energy growth (due to the underlying non-normality of the dynamics) before settling onto a (forced) time-periodic limit cycle.
This behavior is clearly visible in Fig.~\ref{fig:cases}, where the energy $E(t)$ will be defined later.
The three training/testing regimes are discussed below, with training happening over the temporal interval $[t_0, t_1]$ and testing over $(t_1, t_2]$.

\begin{enumerate}
  \item \textbf{Case~1 (rich training):} the final offline time $t_1$ is well inside the oscillatory
        regime, providing a rich dataset that already contains most of the information of the post-transient dynamics (see Fig.~\ref{subfig:case1}).
  \item \textbf{Case~2 (regime change):} the final offline time $t_1$ falls at the very beginning of the post-transient phase, requiring the ROM to adapt to new oscillatory patterns
        during online prediction (see Fig.~\ref{subfig:case2}). \reviewerone{Here, ``regime change'' denotes the transition from transient to sustained oscillatory
dynamics that are absent from the offline training window.}
  \item \textbf{Case~3 (minimal training):} $t_1$ is before the transient onset,
        corresponding to extremely limited and low-energy training data,
        where adaptation must recover unseen dynamics from scratch (see Fig.~\ref{subfig:case3}).
\end{enumerate}

\begin{remark}
\reviewertwo{The final test time $t_2$ is chosen separately for each case to maintain a comparable ratio
between the online prediction phase and the available offline training data. In cases~2 and~3, the offline
training window ends significantly earlier, and fixing $t_2$ to the value used in case~1 would
impose a disproportionately longer extrapolation horizon. Moreover, since online adaptation relies on FOM queries initialized from lifted ROM predictions, prediction errors inevitably accumulate over
long horizons, particularly in challenging regimes. Accordingly, we report results over
shorter but still highly challenging prediction intervals in these cases.}
\end{remark}

It is reasonable to question whether these tests are fair for static ROMs, which, by definition, rely on a larger corpus of training data. However, our experimentations with a wider range of training data on more complex problems (e.g. ~\cite{huang2022model,arnold2022large}) has shown that these test cases are representative of challenges faced by static ROMs in settings with strong transients, transport-dominated physics, and particularly in chaotic problems where every snapshot is unique, and out-of-manifold.

\subsection{Offline/static ROM training}
\begin{figure}[t]
  \centering
  \includegraphics[width=0.4\linewidth]{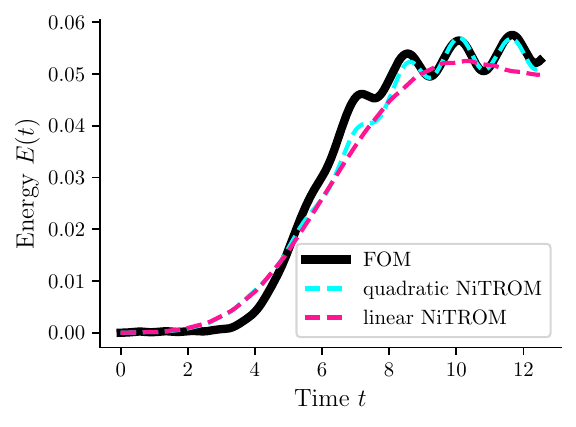}
  \caption{{Performance of $10$-dimensional linear vs. quadratic static NiTROM models against the training trajectory. We notice that the linear model is unable to track the oscillatory motion of the energy, while the quadratic ROM captures the true trajectory.}}
  \label{fig:linear-vs-quadratic}
\end{figure}

\begin{figure}[htb!]
    \centering
    \begin{subfigure}[t]{0.49\textwidth}
        \includegraphics[width=\textwidth]{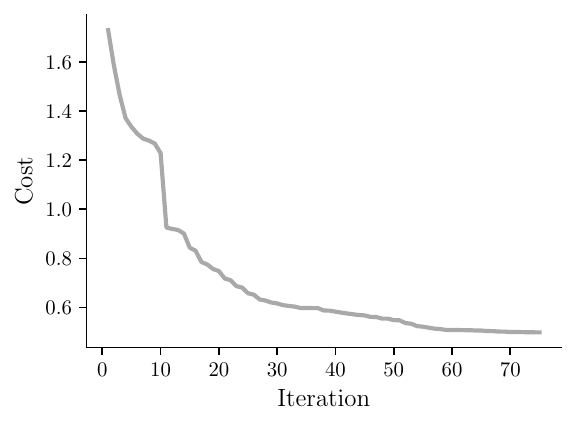}
        \vspace{-2em}
        \caption{{Case~1 (rich training)}}
    \end{subfigure}
    \hfill
    \begin{subfigure}[t]{0.49\textwidth}
        \includegraphics[width=\textwidth]{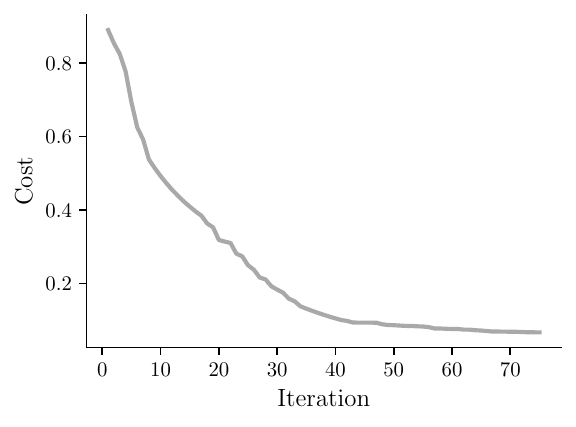}
        \vspace{-2em}
        \caption{{Case~2 (regime change)}}
    \end{subfigure}
    % \hspace{-1.em}
    \begin{subfigure}[t]{0.49\textwidth}
        \includegraphics[width=\textwidth]{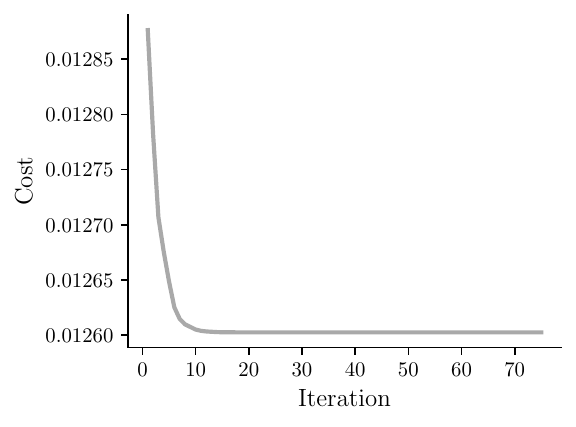}
        \vspace{-2em}
        \caption{{Case~3 (minimal training)}}
    \end{subfigure}
    
    \caption{\reviewerthree{Convergence histories of the offline NiTROM optimization used to train the static ROM
for each case.}}
    \label{fig:nitrom_offline_convergence}
\end{figure}

Static models (Galerkin, OpInf, and NiTROM)
are trained on the snapshots from $[t_0,t_1]$.
\reviewerthree{The static Galerkin ROM is constructed using the same POD basis as the static OpInf model.}
In all cases, we design a quadratic ROM as given
in Eq.~\eqref{eq:poly_reduced}, since a purely linear formulation was found
to be too restrictive. 
In fact, it fails to reproduce the cyclic energy
behavior observed during training, whereas the quadratic model
accurately captures these in-distribution oscillations (Fig.~\ref{fig:linear-vs-quadratic}). \reviewerthree{Indeed, the presence of quadratic
terms in the reduced dynamics is physically expected, since the governing Navier--Stokes
equations in Eq.~\eqref{eq:ns_fom} contain a quadratic convection term.}
The reduced dimension $r$ is chosen as $r=10$, which captures the
dominant energetic modes without excessive computational cost.
While in classical static ROMs, $r$ is often selected based on cumulative
POD energy content, this restriction is less critical in the adaptive
context because the model parameters evolve over time and can gradually
absorb new information.
\reviewerthree{In the case of static NiTROM, a fixed number of offline optimization steps is taken over
the manifold to ensure accurate modeling of the training trajectory.
Figure~\ref{fig:nitrom_offline_convergence} shows the convergence behavior of the offline NiTROM optimization for all three
cases considered in this work. In case~1, convergence is achieved within approximately
60 iterations, while case~2 requires 50 iterations and case~3 converges after only 10 steps. These three static NiTROM models are then used as the initial condition for corresponding adaptive ROMs in each case.}

\subsection{Online/adaptive ROM setup}
During the online phase over the testing window $(t_1,t_2]$, we activate the adaptive mechanisms
described in Sec.~\ref{subsec:adaptive_philosophy} and Alg.~\ref{alg:unified_framework}.
Between two adaptation events, the ROM propagates forward for $Z$
time steps. Then, the adaptation is triggered and the model
interacts with the FOM to obtain a new snapshot before updating
its parameters using the data window $\mathcal{D}_j$ of length~$M$.
We perform ablation studies by sweeping over the main hyperparameters
$(M,Z,K)$ to assess the impact of adaptation frequency, window length,
and optimization budget on stability and predictive accuracy.

\subsection{Evaluation metrics}
\label{subsec:metrics}

The predictive quality of each ROM is assessed using both quantitative
and qualitative measures.
The primary diagnostic is the perturbation energy,
\[
  E(t) = \|\mathbf{v}\|_2^2,
\]
which directly measures the amplitude of the velocity fluctuations $\vecv$ around the underlying steady-state velocity profile $\overline{\mathbf{V}}$ shown in Fig.~\ref{fig:FOM}a.
In addition to energy, we also compute the absolute field error
\[
  e(t) = \|\mathbf{v}_{\mathrm{FOM}}(t)
          - \mathbf{v}_{\mathrm{ROM}}(t)\|_2
\]
that provides a global measure of accuracy over the spatial
domain.
\reviewerthree{To complement these global metrics, we examine the $u$-velocity profile along a horizontal slice located at
$y = 0.05\,L_y$ (here, $L_y=1$) above the bottom wall, where most of the dominant flow
activity occurs.}
This provides a compact, quantitative
representation of the local amplitude and phase of the
dynamics, closely matching what is observed in the two-dimensional field
plots.
Finally, we visually inspect the global vorticity and velocity fields to whether the phase and amplitude of the predicted flow structures match those of the full-order model.

%%%%%%%%%%%%%%%%%%%%%%%%%%%%%%%%%%%%%%%%%%%%%%%%%%%%%%%%%%%%%
\section{Results \& Discussion}
\label{sec:results}
%============================================================
\subsection{Case 1: Rich training}
\label{subsec:case1}
%============================================================

In this case, the models are trained over $[0,12.5]$ and then used
to predict the subsequent $(12.5,20]$ interval (Fig.~\ref{subfig:case1}).
The training window contains the entire transient and several
oscillatory periods of the flow, meaning that most relevant spatial and
temporal structures have already been observed.
This makes the online stage relatively simple, where only small amplitude
and phase corrections are needed.

\begin{figure}[t]
  \centering
  \includegraphics[width=\linewidth]{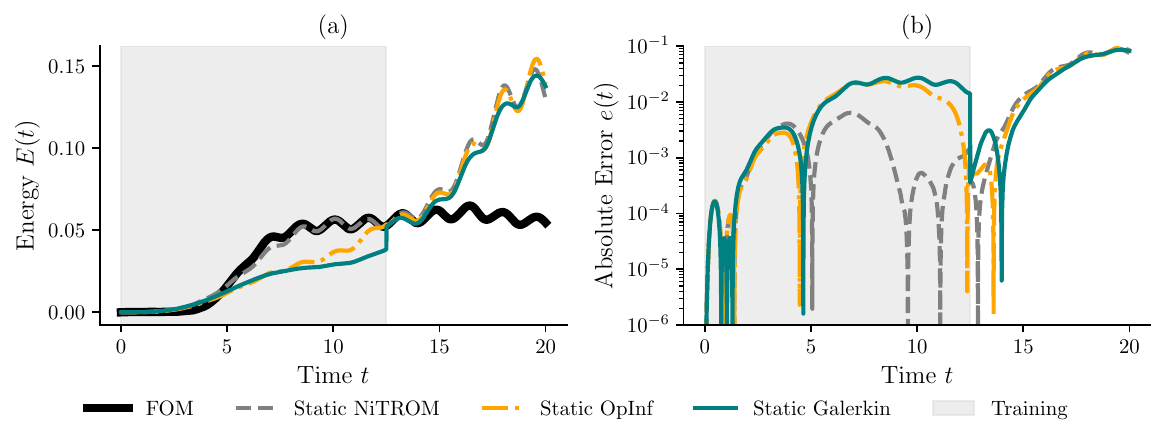}
  \caption{{Energy and field-error evolution for static ROMs in case 1 (rich training). 
All static models (Galerkin, OpInf, NiTROM) match the general oscillatory pattern but gradually overpredict energy, 
leading to growing amplitude and loss of stability beyond the training horizon. Note that NiTROM performs much better compared to the other static models inside the training window.}}
  \label{fig:case1_energy_error}
\end{figure}

\subsubsection{Static ROMs}
Figure~\ref{fig:case1_energy_error} shows the evolution of the energy
and the corresponding field error for the static Galerkin, OpInf, and
NiTROM models.
All three models reproduce the general oscillatory pattern of
the FOM but quickly drift away, indicating
amplitude overgrowth that would likely lead to instability if the
integration were continued.
\reviewerthree{Note that Galerkin projection intrusively derives the reduced operators by projecting the
full-order governing equations onto this basis, while OpInf identifies the reduced dynamics from
data via regression, leading to different reduced operators and, in general, different predictive
behavior despite the shared basis.}
Within the training window, static NiTROM achieves the best fit to the
data, reflecting the benefit of its offline manifold optimization and
oblique projection.
However, in the prediction window, all static models behave comparably.
This shows that the extrapolation accuracy achieved through offline optimization is
limited, and an online adaptation mechanism is required to maintain
accuracy.
Inspection of the vorticity fields in Fig.~\ref{fig:case1_vor} at the end of the
prediction horizon confirms this trend.
While the dominant flow structures and phase patterns are captured
correctly, the predicted fields exhibit visibly larger amplitudes than the fields from the FOM, and this is consistent with the observed energy overshoot.
Appendix~\ref{app:fields} contains (perturbed) velocity component fields $u$ and $v$ for all methods
tested in this paper, across all cases.

\subsubsection{Adaptive ROMs at frequent updates}
\begin{figure}[t]
  \centering
  \includegraphics[width=\linewidth]{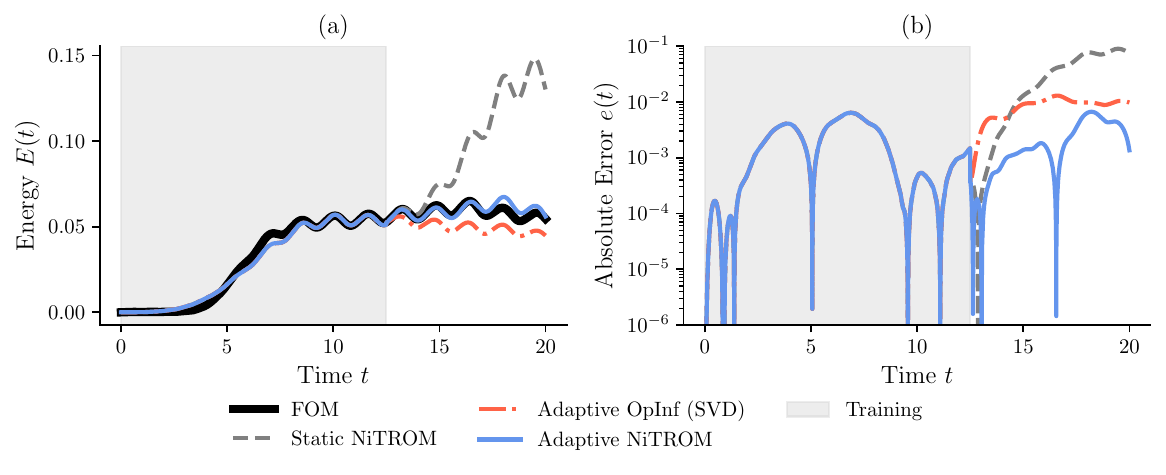}
  \caption{{Energy and field-error evolution for adaptive ROMs ($Z=10$, $M=100$) in case 1 (rich training). 
Both adaptive models suppress the nonphysical energy growth of static ROMs. 
Adaptive OpInf exhibits mild damping, while Adaptive NiTROM (with $K=10$) nearly matches the FOM energy.}}
  \label{fig:case1_z10_energy}
\end{figure}

\begin{figure}[t]
  \centering
  \includegraphics[width=0.6\linewidth]{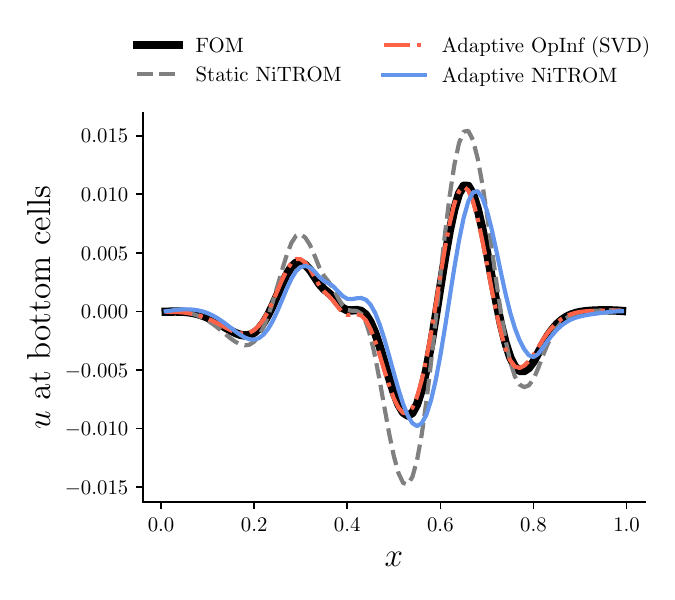}
  \caption{{$u$-velocity slice for adaptive ROMs ($Z=10$, $M=100$) at a horizontal location
$y = 0.05\,L_y$ above the bottom wall in case 1 (rich training) at $t=20$. 
Both Adaptive OpInf and Adaptive NiTROM accurately reproduce the amplitude of the near-wall velocity, 
confirming their ability to control energy growth while maintaining physical coherence. Adaptive OpInf also captures the phase correctly, while Adaptive NiTROM causes a small phase shift.}}
  \label{fig:case1_z10_u-slice}
\end{figure}

\begin{figure}[t]
  \centering
  \includegraphics[width=\linewidth]{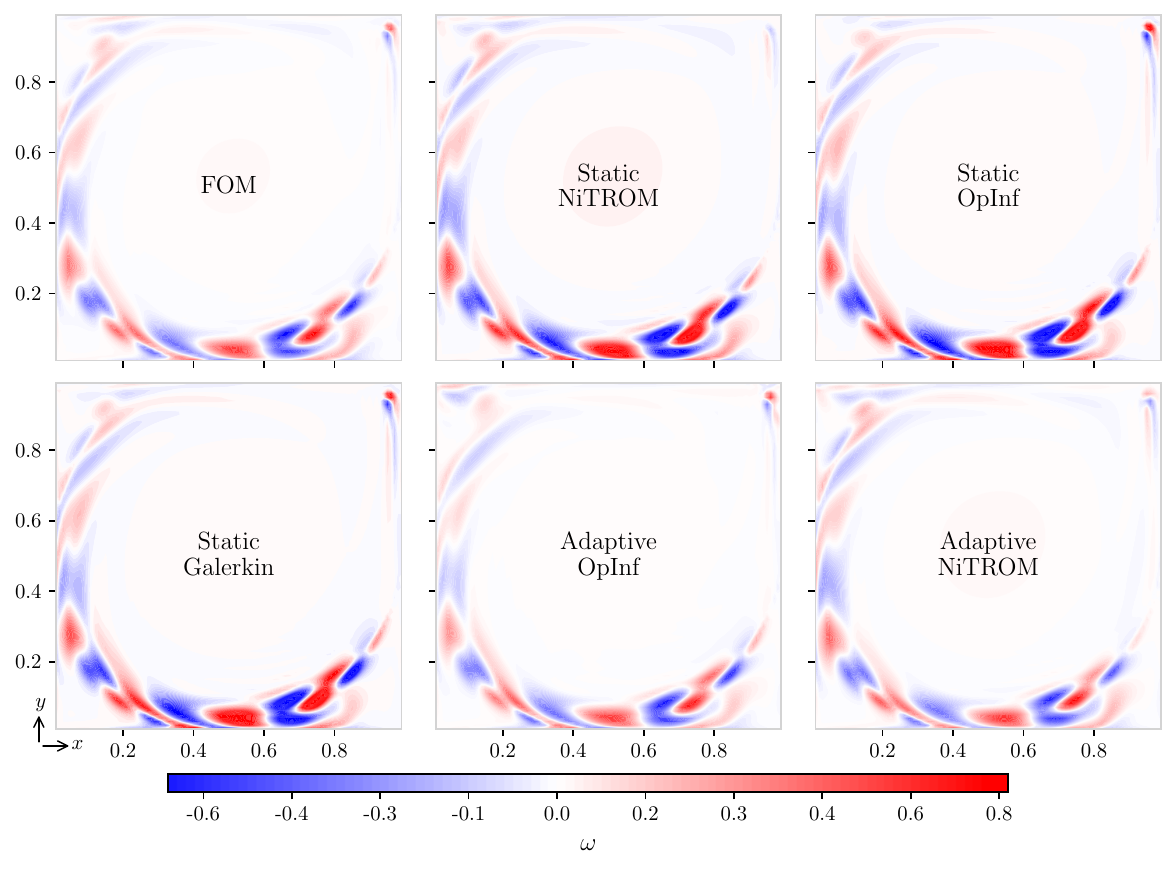}
  \caption{Ground truth (FOM) and predicted (ROM) vorticity fields at the end of the online window ($t=20$) in case 1. All models are quadratic and $r=10$-dimensional. For adaptive models we have an adaptation window $Z=10$, a lookback window $M=100$, and a per-adaptation optimization budget $K=10$.}
  \label{fig:case1_vor}
\end{figure}

Next, we examine the adaptive models with an adaptation window of
$Z=10$ and a lookback window of $M=100$ snapshots.
For Adaptive OpInf, the basis is updated using windowed SVD, and for
Adaptive NiTROM, $K=10$ optimization iterations are performed per update.
Looking at the energy and error plots in Fig.~\ref{fig:case1_z10_energy}, both approaches successfully suppress the non-physical energy growth
observed in the static models.
Adaptive OpInf shows a slight decay in energy, suggesting mild
over-damping that could lead to amplitude fading over very long
horizons.
Adaptive NiTROM, in contrast, almost perfectly matches the FOM energy
curve.
Velocity slice (Fig.~\ref{fig:case1_z10_u-slice}) and vorticity field (Fig.~\ref{fig:case1_vor}) comparisons reveal that both adaptive
models reproduce the dynamics accurately and maintain amplitude control,
while Adaptive NiTROM introduces a small phase lag relative to the FOM.
This lag could be mitigated in principle by augmenting NiTROM’s loss
function with a phase-alignment term.

\subsubsection{Adaptive ROMs at less frequent updates}
\begin{figure}[t]
  \centering
  \includegraphics[width=\linewidth]{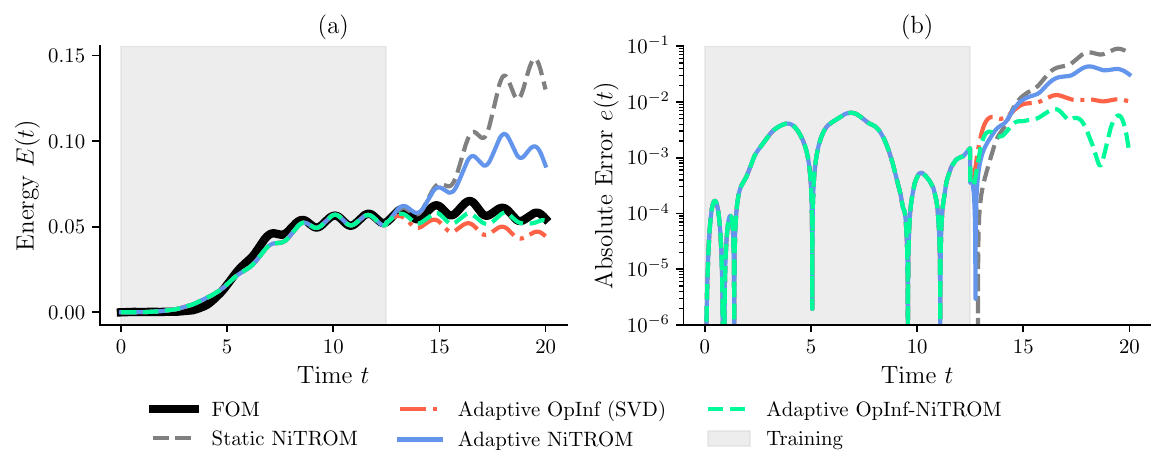}
  \caption{{Energy and field-error evolution for adaptive ROMs ($Z=50$, $M=20$) in case 1 (rich training). 
Adaptive OpInf continues to stabilize the solution but slightly over-damps energy. 
Adaptive NiTROM (with $K=20$) struggles due to its warm-start sensitivity, while the Adaptive OpInf--NiTROM (with $K=10$) restores near-perfect energy tracking.}}
  \label{fig:case1_z50_energy}
\end{figure}

\begin{figure}[t]
  \centering
  \includegraphics[width=0.6\linewidth]{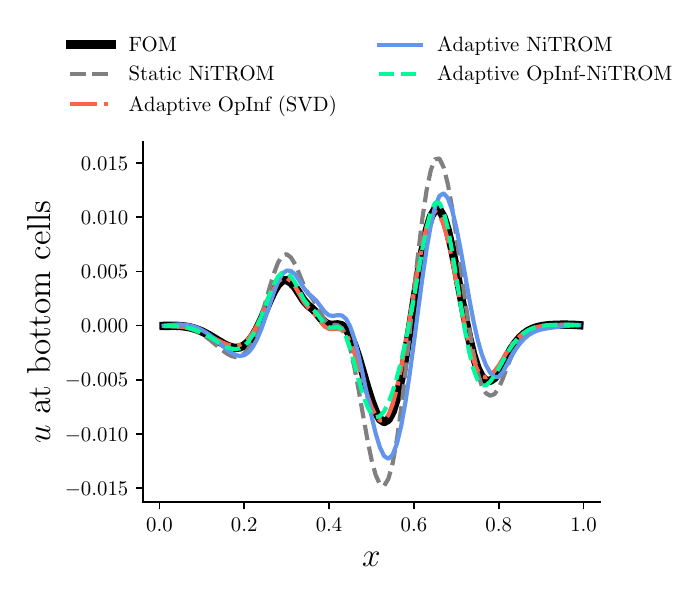}
  \caption{{$u$-velocity slice for adaptive ROMs ($Z=50$, $M=20$) at a horizontal location
$y = 0.05\,L_y$ above the bottom wall in case 1 (rich training) at $t=20$. 
The hybrid model preserves both phase and amplitude of the oscillations, similar to Adaptive OpInf. Adaptive NiTROM causes a small phase shift and amplitude growth.}}
  \label{fig:case1_z50_u-slice}
\end{figure}

\begin{figure}[t]
  \centering
  \includegraphics[width=\linewidth]{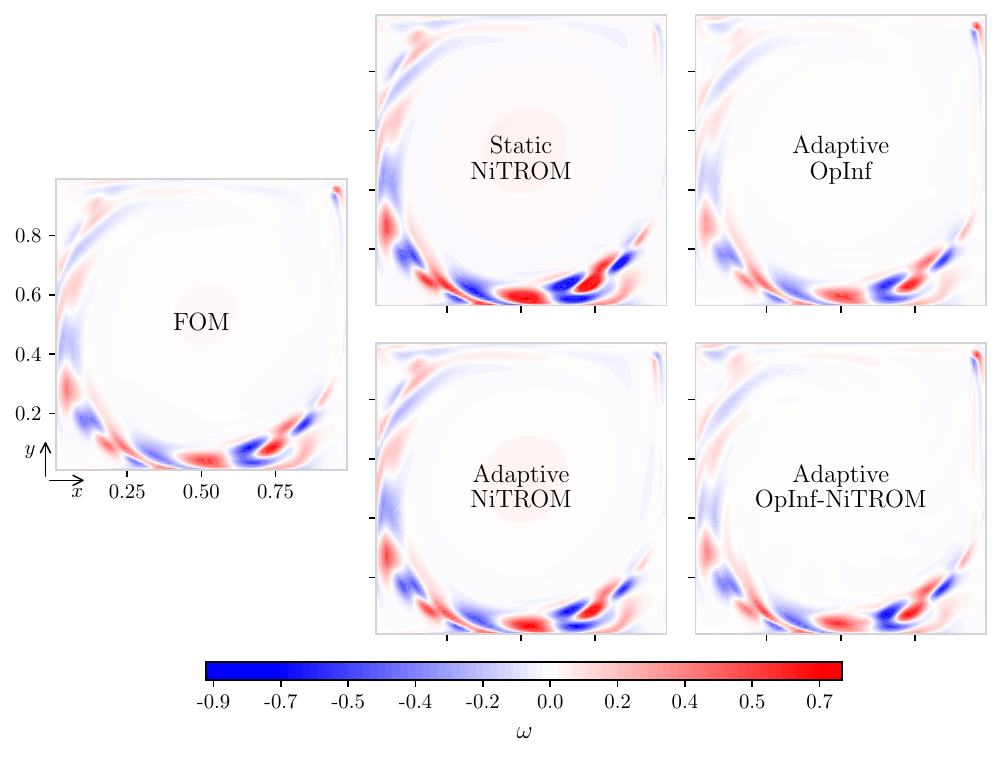}
  \caption{Ground truth (FOM) and predicted (ROM) vorticity fields at the end of the online window ($t=20$) in case 1. All models are quadratic and $r=10$-dimensional. For adaptive models we have an adaptation window $Z=50$ and a lookback window $M=20$. For Adaptive NiTROM and Adaptive OpInf--NiTROM, per-adaptation optimization budget is set to $K=20$ and $K=10$, respectively.}
  \label{fig:case1_z50_vor}
\end{figure}

To investigate the effect of less frequent adaptation, we increased the
adaptation window to $Z=50$ and reduced the lookback to $M=20$.
The energy and error plots are shown in Fig.~\ref{fig:case1_z50_energy}.
In this configuration, Adaptive OpInf continues to control the energy
growth, though, just like the previous case, the overall amplitude decreases
over time.
Adaptive NiTROM, now performing $K=20$ optimization steps per update,
struggles to recover the correct energy trajectory.
This is due to its optimization nature and its warm-start initialization
from the previously learned parameters.
When the system evolves substantially between two adaptation events, the
new optimal parameters may lie far from the previous ones in the
optimization landscape, causing the algorithm to converge slowly or fall
into local minima.
This observation highlights a key limitation of purely optimization-based adaptation.
To address this issue, we tested our Adaptive OpInf--NiTROM method that performs a quick
OpInf refit followed by $K=10$ NiTROM optimization steps.
As expected, this configuration performs better than
either adaptive method alone.
The OpInf stage rapidly re-learns the operators based on the current
window, effectively providing a good initialization, while the NiTROM
refinement fine-tunes the parameters over the manifold.
The resulting energy trajectory in Fig.~\ref{fig:case1_z50_energy}
aligns closely with the FOM, confirming
that such hybrid adaptation can effectively handle situations where the
system changes substantially between updates.
Nevertheless, this configuration does not fully showcase the potential of the hybrid method.
As evident from the velocity slice in Fig.~\ref{fig:case1_z50_u-slice} and qualitative field comparison in Fig.~\ref{fig:case1_z50_vor}, Adaptive OpInf already reproduces the flow structures accurately, leaving little room for improvement through additional manifold-optimization steps, despite the slight gains observed in the energy curve.
The following cases (2 and 3) introduce more challenging conditions where the hybrid method demonstrates its clear advantage over the individual approaches.

\subsubsection{Ablation studies}
\begin{figure}[htb!]
    \centering
    \begin{subfigure}[t]{0.49\textwidth}
        \includegraphics[width=\textwidth]{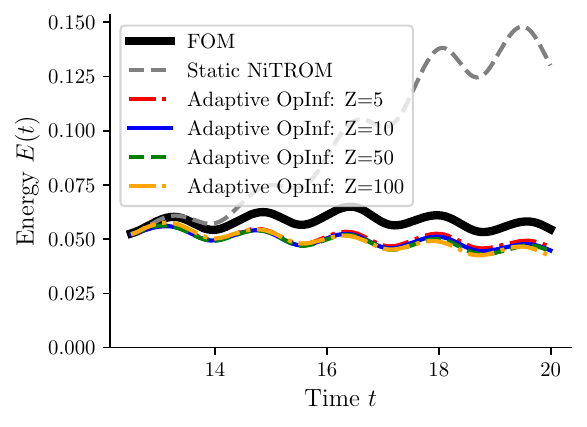}
        \vspace{-2em}
        \caption{{Effect of varying $Z$ (with variable $M$ for each $Z$, to ensure the window spans $1,000$ time steps)}}
        \label{fig:opinf_ablations_Z}
    \end{subfigure}
    \hfill
    \begin{subfigure}[t]{0.49\textwidth}
        \includegraphics[width=\textwidth]{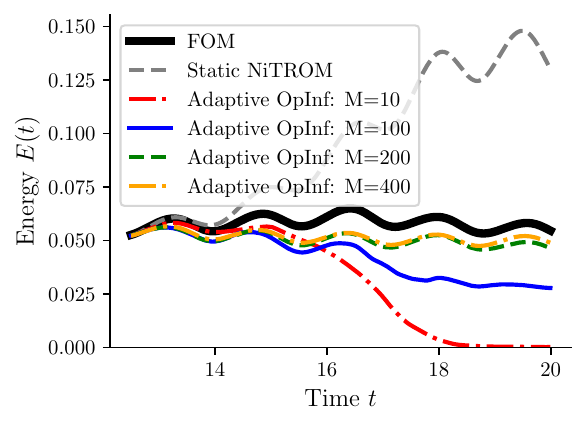}
        \vspace{-2em}
        \caption{{Effect of varying $M$ (with frozen $Z=10$)}}
        \label{fig:opinf_ablations_M}
    \end{subfigure}
    % \hspace{-1.em}
    \begin{subfigure}[t]{0.49\textwidth}
        \includegraphics[width=\textwidth]{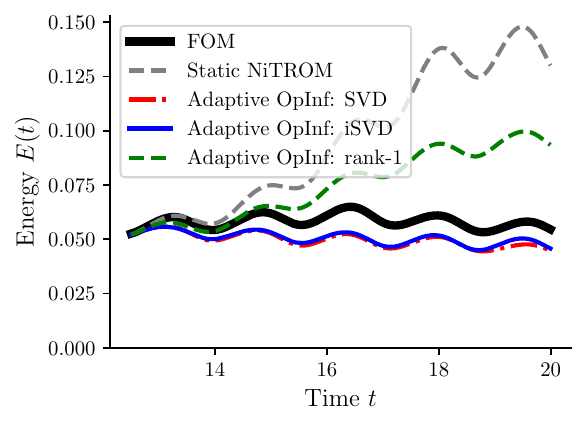}
        \vspace{-2em}
        \caption{{Effect of basis adaptation method (with frozen $Z=10$ and $M=100$)}}
        \label{fig:opinf_ablations_basis}
    \end{subfigure}
    
    \caption{{Ablation studies for Adaptive OpInf applied to case~1. Here, $Z$ and $M$ denote, respectively, adaptation window and lookback window. Shown are performances over the online (prediction) window only, as training is identical for all models.}}
    \label{fig:opinf_ablations}
\end{figure}

\begin{figure}[htb!]
    \centering
    \begin{subfigure}[t]{0.49\textwidth}
        \includegraphics[width=\textwidth]{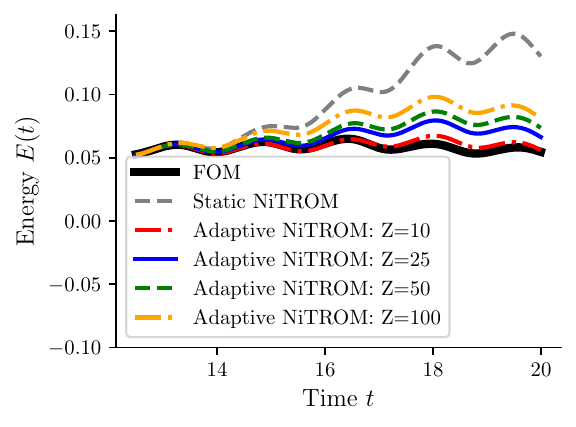}
        \vspace{-2em}
        \caption{{Effect of varying $Z$ (with frozen $K=10$ and variable $M$ for each $Z$, to ensure the window spans $1,000$ time steps)}}
        \label{fig:nitrom_ablations_Z}
    \end{subfigure}
    \hfill
    \begin{subfigure}[t]{0.49\textwidth}
        \includegraphics[width=\textwidth]{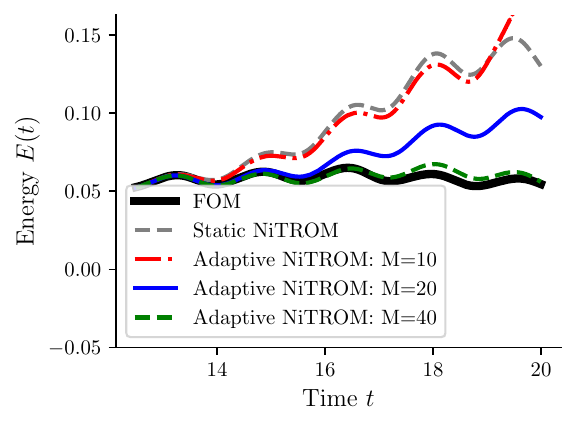}
        \vspace{-2em}
        \caption{{Effect of varying $M$ (with frozen $Z=10$ and $K=10$). \reviewertwo{Note that NiTROM converges to the ground truth with significantly smaller window sizes compared to OpInf.}}}
        \label{fig:nitrom_ablations_M}
    \end{subfigure}
    % \hspace{-1.em}
    \begin{subfigure}[t]{0.49\textwidth}
        \includegraphics[width=\textwidth]{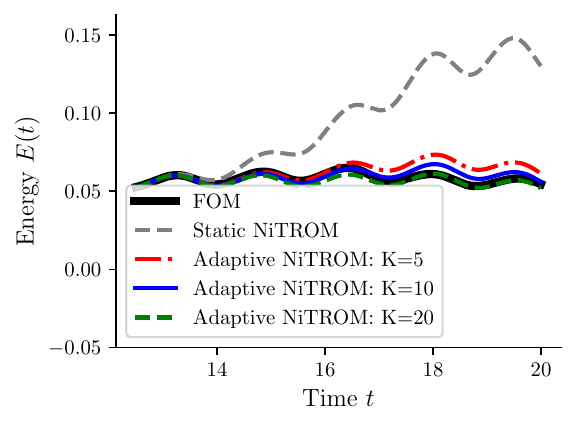}
        \vspace{-2em}
        \caption{{Effect of varying $K$ (with frozen $Z=10$ and $M=40$)}}
        \label{fig:nitrom_ablations_K1}
    \end{subfigure}
    \hfill
    \begin{subfigure}[t]{0.49\textwidth}
        \includegraphics[width=\textwidth]{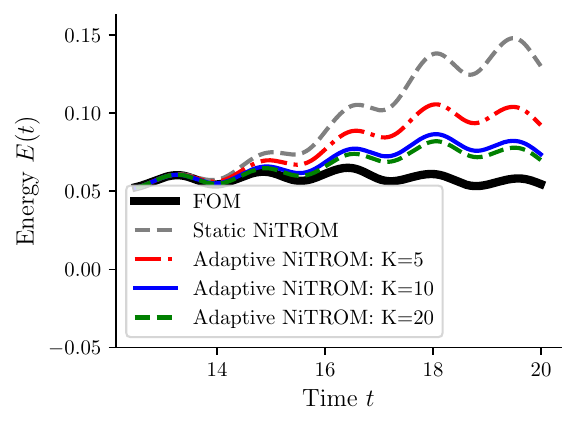}
        \vspace{-2em}
        \caption{{Effect of varying $K$ (with frozen $Z=50$ and $M=10$)}}
        \label{fig:nitrom_ablations_K2}
    \end{subfigure}
    
    \caption{{Ablation studies for Adaptive NiTROM applied to case~1. Here, $Z$, $M$, and $K$ denote, respectively, adaptation window, lookback window, and per-adaptation manifold optimization steps. Shown are performances over the online (prediction) window only, as training is identical for all models.}}
    \label{fig:nitrom_ablations}
\end{figure}

\begin{figure}[t]
  \centering
  \includegraphics[width=0.49\linewidth]{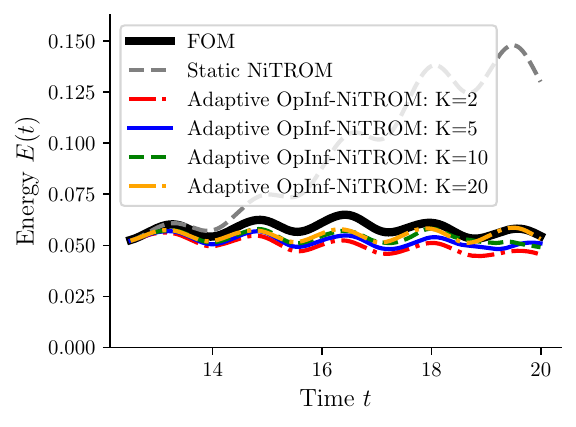}
  \caption{Effect of number of per-adaptation manifold optimization steps $K$ for Adaptive (hybrid) OpInf--NiTROM applied to Case~1, with frozen $Z=50$ and $M=20$. Shown are performances over the online (prediction) window only, as training is identical for all models.}
  \label{fig:hybrid_ablations}
\end{figure}

A detailed ablation study was performed to assess the sensitivity of the
adaptive models to their hyperparameters and design choices.
The complete set of experiments is presented in Figs.~\ref{fig:opinf_ablations}--\ref{fig:hybrid_ablations}.
The following conclusions summarize the main findings:
\begin{itemize}
  \item \textbf{Effect of adaptation window ($Z$):}  
        Adaptive OpInf remains robust over a wide range of $Z$ values and
        maintains stable predictions even with large update intervals (see Fig.~\ref{fig:opinf_ablations_Z}).  
        Adaptive NiTROM performs best for small $Z$, where the optimizer
        starts close to the current minimum, and its accuracy degrades
        as $Z$ increases (see Fig.~\ref{fig:nitrom_ablations_Z}). Note that in Figs.~\ref{fig:opinf_ablations_Z} and~\ref{fig:nitrom_ablations_Z},
        the window size
        $M$ is chosen for each value of $Z$ such that all models span the same temporal interval. This
        ensures that observed differences are not influenced by mismatched lookback horizons; further
        details are discussed below.
  \item \textbf{Effect of lookback window ($M$):}  
        The lookback window should be chosen consistently with $Z$.
        Once $Z$ is chosen, the quantity $(M-1)Z\Delta t$ effectively determines the temporal span of the system’s past history considered during each adaptation (Fig.~\ref{fig:adaptation_schematic}). The appropriate choice of $M$ is therefore problem-dependent and should be informed by the characteristic physical timescales of the underlying dynamics. For instance, we see in Figs.~\ref{fig:opinf_ablations_M} and~\ref{fig:nitrom_ablations_M} that a small $M$ fails to capture enough temporal variation for effective learning.
  \item \textbf{Basis-update strategy:}
        In this work, we employ the windowed SVD as the default basis update strategy, as it provides a clear and consistent baseline that does not rely on additional algorithmic choices.
        The ablation results in Fig.~\ref{fig:opinf_ablations_basis}, however, suggest that iSVD may offer a promising direction for future exploration,
        since it delivers similar performance to the windowed SVD in case 1, while incurring substantially lower computational cost.
        As discussed earlier, rank-one basis updates were found to be ineffective in this non-intrusive setting, where local intrusive modifications to the basis are not possible.
  \item \textbf{Optimization steps ($K$) in NiTROM:}  
        Increasing $K$ generally improves adaptation accuracy (as seen in Fig.~\ref{fig:hybrid_ablations}), but
        overly large values add
        unnecessary computational burden.
        \reviewerthree{In regimes where the dynamics evolve rapidly, excessively large values of $K$ may reduce
        responsiveness by over-emphasizing data from a local window that is no longer fully
        representative of the current dynamics, whereas in slowly varying regimes this effect is not
        observed.}

\end{itemize}

%============================================================
\subsection{Case 2: Regime change}
\label{subsec:case2}
%============================================================

We tested the model in another setting, where the offline window spans from $t=(0,7.5)$ and the online
prediction covers the following $t=(7.5,10)$, corresponding to $1000$ extrapolation steps (Fig.~\ref{subfig:case2}).
Thus, the models are trained only on the low-energy and transient-growth
phases and must extrapolate into the subsequent oscillatory regime.
For the adaptive runs we fix the adaptation window to $Z=10$, the lookback window
to $M=10$, and (for NiTROM-based updates) the optimization budget to
$K=10$ iterations per adaptation.
Adaptive OpInf uses windowed SVD for the basis update.

%---------------------------- Static ----------------------------

\subsubsection{Static ROMs}
\begin{figure}[t]
  \centering
  \includegraphics[width=\linewidth]{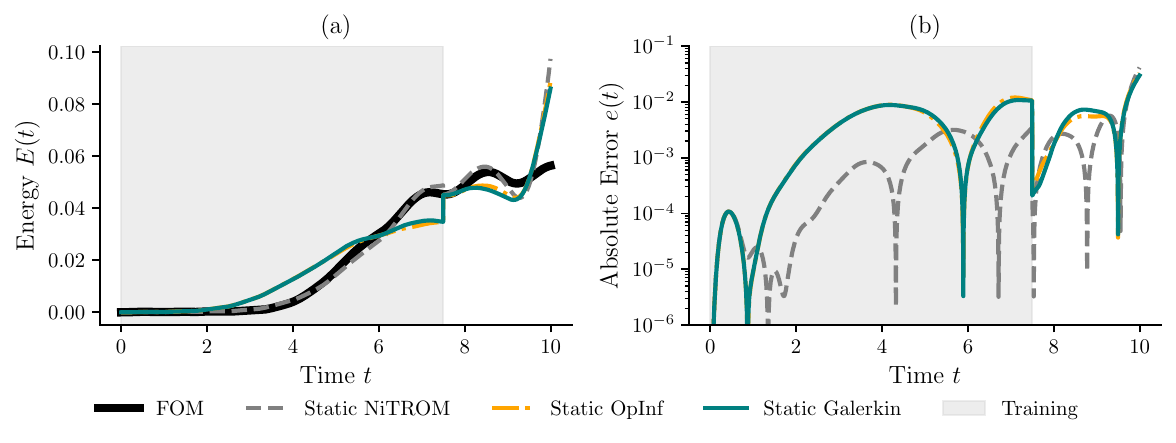}
  \caption{{Energy and field-error evolution for static ROMs in case 2 (regime change).
  All static models fail to reproduce the regime change. 
While static NiTROM captures the first oscillation, it soon diverges, highlighting the inability of static ROMs to generalize to unseen dynamical regimes.}}
  \label{fig:case2_energy_static}
\end{figure}

Figure~\ref{fig:case2_energy_static} shows that all static ROMs
fail to reproduce the new oscillatory regime.
Static Galerkin and static OpInf diverge rapidly in energy.
Static NiTROM tracks the very first oscillation,
but then the energy increases uncontrollably and the solution departs
from the physical attractor, similar to the other static models.
Looking at the predicted fields at the end of the online phase for
Adaptive NiTROM in Fig.~\ref{fig:case2_vor},
we see that a static model is unable to faithfully represent the true dynamics,
confirming the non-physical nature of the static
predictions once the system leaves the training distribution.

%---------------------------- Adaptive ----------------------------

\subsubsection{Adaptive ROMs}
\begin{figure}[t]
  \centering
  \includegraphics[width=\linewidth]{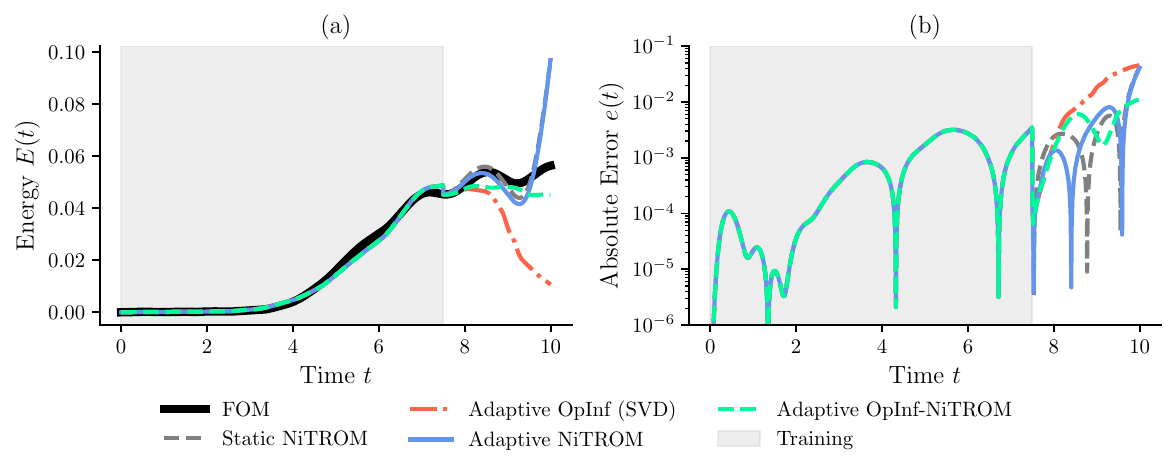}
  \caption{{Energy and field-error evolution for adaptive ROMs ($Z=10$, $M=10$) in case 2 (regime change).
  Adaptive OpInf stabilizes the trajectory but underpredicts the oscillation amplitude. 
Adaptive NiTROM (with $K=10$) adds no improvement over static NiTROM, whereas the Adaptive OpInf--NiTROM (with $K=10$) approach maintains a bounded, physically consistent energy evolution.}}
  \label{fig:case2_energy}
\end{figure}

With $Z=10$ and $M=10$, the Adaptive OpInf curve in
Fig.~\ref{fig:case2_energy} shows a clear energy drop
in the online window, showing that dynamics gradually fade away.
While this stabilizes the solution, it underestimates the amplitude of
the true oscillations and therefore does not capture the regime change
quantitatively.
Adaptive NiTROM, using $K=10$ iterations per update, yields essentially
no improvement over static NiTROM, as the energy grows similarly after the
first cycle.
This mirrors the behavior observed in case 1 with sparse updates and is
consistent with the mechanism of NiTROM’s warm-started manifold
optimization.

In contrast, the Adaptive OpInf--NiTROM configuration retains
the overall energy level over the prediction window.
Although the hybrid curve does not perfectly match the detailed
oscillatory behavior, the total energy is maintained rather than
damped (as in Adaptive OpInf) or amplified (as in Adaptive NiTROM).

\begin{figure}[t]
  \centering
  \includegraphics[width=0.6\linewidth]{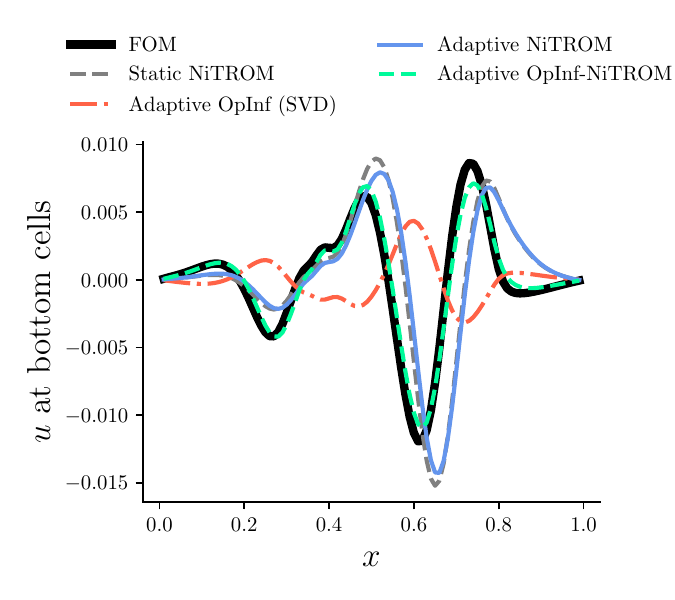}
  \caption{{$u$-velocity slice for adaptive ROMs ($Z=10$, $M=10$) at a horizontal location
$y = 0.05\,L_y$ above the bottom wall in case 2 (regime change) at $t=10$.
  Adaptive OpInf fails to capture the true profile, Adaptive NiTROM exhibits minimal corrections compared to its static counterpart, while the hybrid method aligns amplitude and phase with the FOM.}}
  \label{fig:case2_u-slice}
\end{figure}

\begin{figure}[t]
  \centering
  \includegraphics[width=\linewidth]{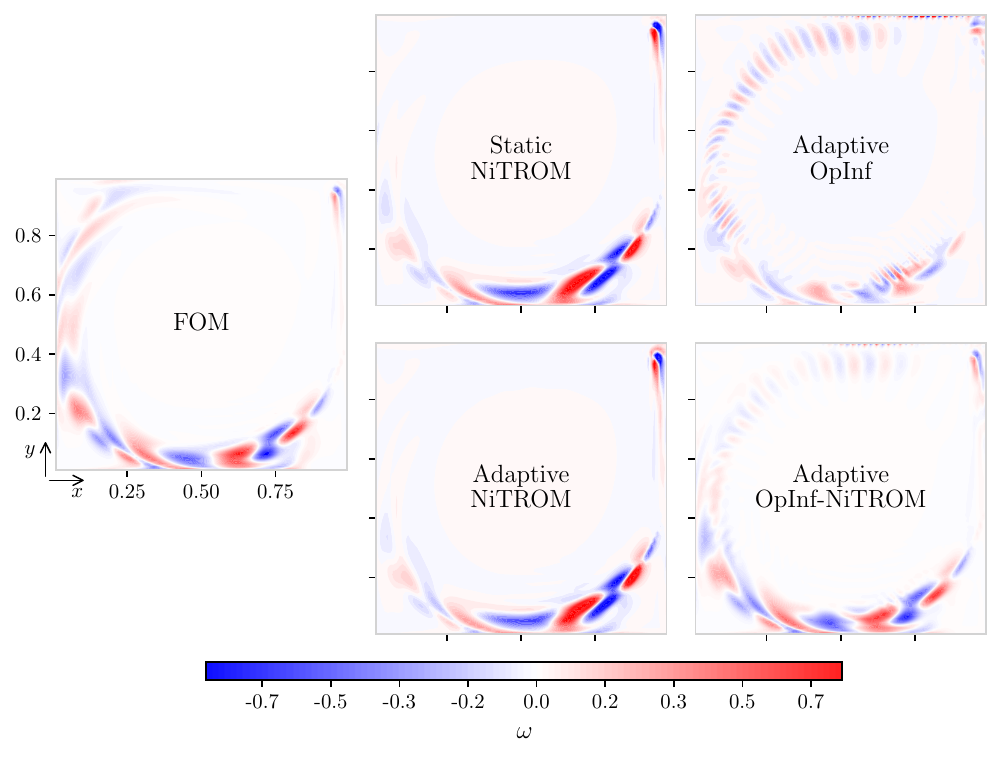}
  \caption{Ground truth (FOM) and predicted (ROM) vorticity fields at the end of the online window ($t=10$) in case 2. All models are quadratic and $r=10$-dimensional. For adaptive models we have an adaptation window $Z=10$, a lookback window $M=10$, and a per-adaptation optimization budget $K=10$.}
  \label{fig:case2_vor}
\end{figure}

Velocity slices (Fig.~\ref{fig:case2_u-slice}) and field visualizations (Fig.~\ref{fig:case2_vor}) clarify these
differences.
Adaptive OpInf,
while suppressing energy, introduces noticeable phase error in the oscillatory
motion and exhibits small numerical wiggles in portions of the domain.
The considerable phase shift is also visible in the $u$-slice.
We also see that Adaptive NiTROM alone adds no visible improvement over the
static counterpart, matching our observation from the energy plot.

The hybrid method yields the most compelling fields.
We observe clean vortical structures with accurate placement and
strength, and the $u$-slice shows good amplitude control and phase
alignment relative to the FOM.
Importantly, the small oscillations present in Adaptive OpInf predictions are
substantially reduced after the brief NiTROM refinement, though faint remnants persist in the same
regions.
This is a representative case where purely numerical metrics
can be misleading. Despite the hybrid energy curve not perfectly
tracing the oscillatory motion, the qualitative field assessment shows that
the hybrid model reconstructs the regime change most faithfully.

%============================================================
\subsection{Case 3: Minimal training}
\label{subsec:case3}
%============================================================

In this most challenging configuration, the models are trained on $t=(0,3.75)$
and then tested on $t=(3.75,6.25)$ (Fig.~\ref{subfig:case3}).
The training data correspond to the initial low-energy regime of the
flow, containing almost no dynamic evolution.
Immediately after the training window, the FOM exhibits a sharp energy
growth as the flow transitions into the transient phase.
This setting therefore pushes the limits of any ROM methodology, where
the basis and operators have seen almost no variability, and the
subsequent dynamics evolve rapidly in both amplitude and spatial
structure.
Adaptive parameters are fixed to $Z=10$, $M=10$, and $K=10$.
The smaller lookback window ensures that the model relies only on the
most recent information rather than outdated, non-representative states.

%---------------------------- Static ----------------------------

\subsubsection{Static ROMs}
\begin{figure}[t]
  \centering
  \includegraphics[width=\linewidth]{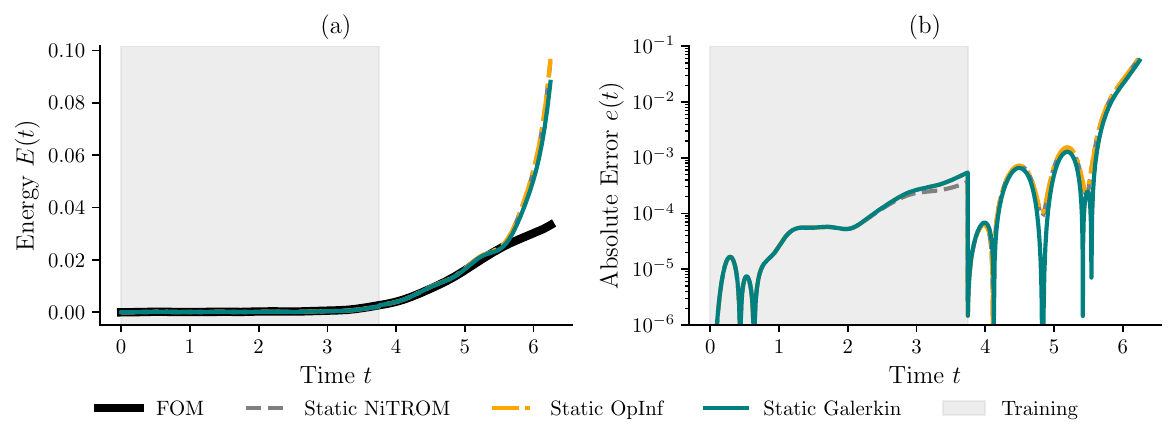}
  \caption{{Energy and field-error evolution for static ROMs in case 3 (minimal training).
  Despite low-energy training data, static ROMs exhibit artificial energy growth.}}
  \label{fig:case3_energy_static}
\end{figure}

Unlike the training data, the static ROMs do not
remain low-energy during prediction.  
Figure~\ref{fig:case3_energy_static} shows that the energy begins to
grow almost immediately in the online phase and diverges from the FOM
towards the end of the prediction window.
Inspection of static NiTROM's flow fields at the end of the online window
(in Fig.~\ref{fig:case3_vor}) reveals that
the predicted dynamics remain localized along the right wall, where the basis was
constructed from the limited low-energy training data.  
Because the basis never observed the broader transient evolution, the
predicted vortices fail to propagate into the domain interior.

%---------------------------- Adaptive ----------------------------

\subsubsection{Adaptive ROMs}

\begin{figure}[t]
  \centering
  \includegraphics[width=\linewidth]{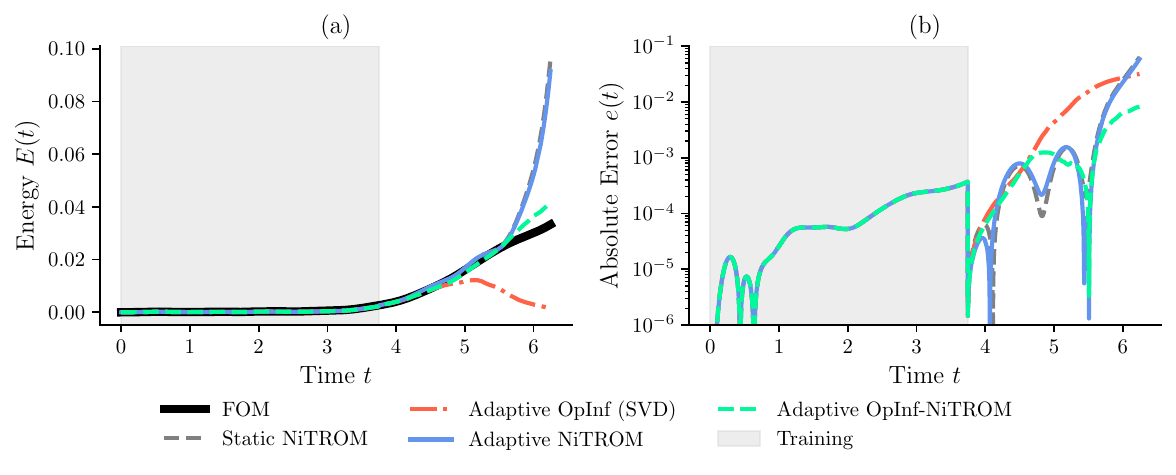}
  \caption{{Energy and field-error evolution for adaptive ROMs ($Z=10$, $M=10$) in case 3 (minimal training).
  Adaptive OpInf stabilizes but underestimates energy; Adaptive NiTROM (with $K=10$) behaves similarly to its static version. 
Adaptive OpInf--NiTROM (with $K=10$) maintains bounded energy and correctly follows the transient growth trend.}}
  \label{fig:case3_energy}
\end{figure}

The adaptive models are also challenged by the rapid changes in this
case.  
As shown in Fig.~\ref{fig:case3_energy}, Adaptive OpInf
produces a noticeable decay in energy, similar to its behavior in
case 2.  
The model remains stable but underestimates the energy amplitude, as the
limited history in the window constrains the temporal context used for
operator inference.  
Adaptive NiTROM, despite its more expressive optimization, again shows
no significant improvement over the static NiTROM.  
Its energy curve follows a similar growth pattern and fails to reproduce
the transient dynamics.

In contrast, the Adaptive OpInf--NiTROM model again achieves
the most stable and physically meaningful results.  
The energy level remains bounded and consistent with the FOM,
neither decaying to zero nor diverging.

\begin{figure}[t]
  \centering
  \includegraphics[width=0.6\linewidth]{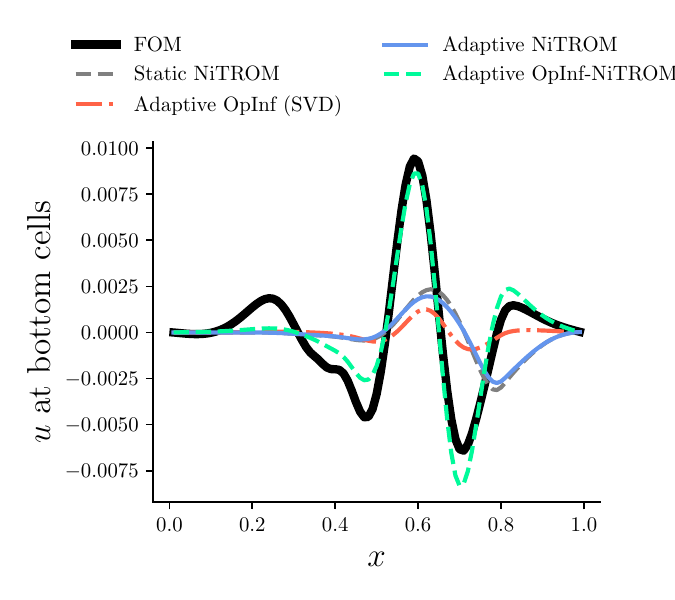}
  \caption{{$u$-velocity slice for adaptive ROMs ($Z=10$, $M=10$) at a horizontal location
$y = 0.05\,L_y$ above the bottom wall in case 3 (minimal training) at $t=6.25$.
  The hybrid method captures both amplitude and phase evolution, reflecting extrapolation capabilities of Adaptive OpInf--NiTROM.}}
  \label{fig:case3_u-slice}
\end{figure}

\begin{figure}[t]
  \centering
  \includegraphics[width=\linewidth]{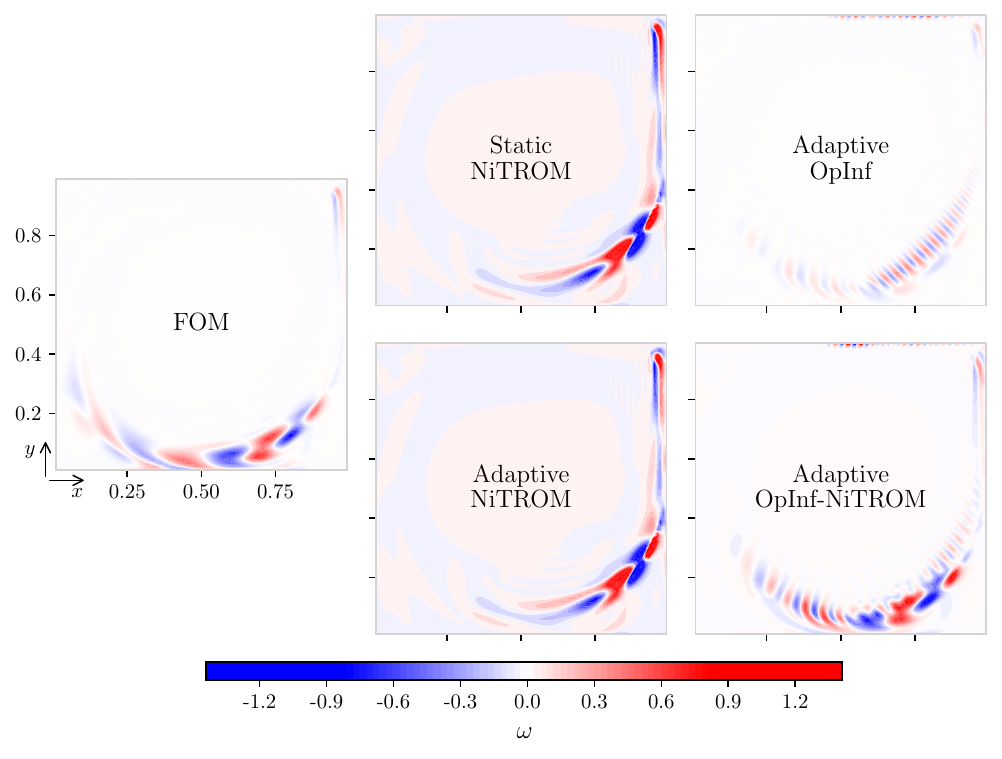}
  \caption{Ground truth (FOM) and predicted (ROM) vorticity fields at the end of the online window ($t=6.25$) in case 3. All models are quadratic and $r=10$-dimensional. For adaptive models we have an adaptation window $Z=10$, a lookback window $M=10$, and a per-adaptation optimization budget $K=10$.}
  \label{fig:case3_vor}
\end{figure}

Now we investigate velocity slice comparison in Fig.~\ref{fig:case3_u-slice}
and field visualizations in Fig.~\ref{fig:case3_vor}
to confirm these quantitative trends.
By close inspection of Adaptive OpInf's predicted fields, we can notice
that it captures only localized amplification near the right wall and fails to reproduce interior vortex propagation.
Adaptive NiTROM, on the other hand, shows no noticeable improvement over its\
static version, consistent with the trends seen in the energy plot.
The Adaptive OpInf--NiTROM approach, however, displays clear
vortex formation and spatial patterns that closely follow the FOM.
The brief NiTROM refinement once again ``cleans up'' the small
oscillatory artifacts produced by the OpInf stage, yielding smoother and
more physically coherent fields.
The $u$-slice captures both the amplitude
growth and the approximate phase of the transient oscillation, showing
that the hybrid method can track the dynamics even when the
offline model has seen virtually nothing.

%============================================================
\subsection{Key observations and takeaways}
\label{subsec:takeaways}
%============================================================

The three cases collectively provide a comprehensive picture of how
different adaptive non-intrusive formulations behave under increasing levels of
difficulty, from minor corrections in a well-trained model to full
extrapolation from nearly featureless data.
Several consistent trends and insights emerge from these experiments.

\paragraph{i. Static ROMs are limited by training coverage.}
Across all cases, static Galerkin, OpInf, and NiTROM
models perform well only when the training window already contains the
dominant flow dynamics.
Outside that regime, they quickly lose physical consistency.
This limitation underscores the need for online correction mechanisms
once the system evolves beyond the distribution represented in the
offline data. 

\paragraph{ii. Adaptive OpInf is robust and efficient.}
In cases with rich training data (case 1), Adaptive OpInf stabilizes the
predictions and prevents long-term energy drift.
In more challenging settings (cases 2 and 3), it is capable of
tracking the qualitative regime change, provided its hyperparameters are
carefully tuned.
Its main strengths are simplicity, low computational cost, and a
remarkable tolerance to large adaptation intervals $Z$.
However, its performance is highly sensitive to the choice of window size $M$, and
it occasionally introduces localized numerical artifacts in the fields.

\paragraph{iii. Adaptive NiTROM is limited by optimization sensitivity and cost.}
The joint manifold optimization in Adaptive NiTROM enables highly
accurate corrections when updates are frequent and the optimization is
well converged (small $Z$, adequate $K$).
In such cases, the energy curve can match the full-order reference
almost perfectly.
However, as the adaptation interval grows, the method
may fail to recover from accumulated error.
In out-of-distribution cases (cases 2 and 3), the optimization
landscape becomes too complex, and the method provides little
benefit over static models.
This sensitivity highlights the difficulty of performing online
trajectory optimization in dynamical systems that change rapidly.
In addition, manifold optimization introduces a significant computational overhead relative to Adaptive OpInf, which limits the suitability of the method for online adaptation. Table~\ref{tab:wallclock} reports the average wall-clock time for a single execution of each operation.

\begin{table}[t]
  \centering
  \caption{Average wall-clock time (in milliseconds) for a single execution of each operation 
  in the adaptive non-intrusive ROM framework. \reviewertwo{All wall-clock times were measured on a local machine equipped with an Apple M3 Pro chip, and are averaged over 10 runs to mitigate measurement noise.} Reported values correspond to a lookback 
  window of $M=10$ snapshots.
  \reviewertwo{Note that methods are initialized from the same offline phase; therefore, the
offline training cost is identical across methods and is not included here.}}
  \label{tab:wallclock}
  \begin{tabular}{lcc}
    \toprule
    \textbf{Operation} & \textbf{Description} & \textbf{Wall-clock time [ms]} \\
    \midrule
    ROM step & one reduced-order model time step & $0.322$ \\
    FOM step & one full-order model time step & $2.33$ \\
    \midrule
    $\textrm{SVD}^\ast$ & one complete SVD over the window & $2.85$ \\
    $\textrm{OpInf}^\ast$ & one least-squares refit of reduced operators & $7.49$ \\
    $\textrm{NiTROM}^\ast$ & one manifold optimization iteration & $31.2$ \\
    \bottomrule
  \end{tabular}
  $^\ast$ represents the fact that these algorithms have not been modified from their offline versions.
\end{table}

\paragraph{iv. Adaptive OpInf--NiTROM combines the strengths of both.}
The Adaptive (hybrid) OpInf--NiTROM approach offers a practical balance
between robustness and accuracy.
The initial OpInf update re-learns the operators from the most recent
data, effectively reinitializing the model for the new regime.
The subsequent NiTROM refinement then enforces geometric consistency and
removes the small numerical artifacts seen in OpInf-only predictions.
Across cases 2 and 3, the hybrid configuration consistently produces
the most physically meaningful and visually coherent fields.
This observation suggests that combining fast regression-based updates with short
manifold refinements is an effective strategy for maintaining accuracy
in challenging settings, although additional work is required to make the
manifold optimization more efficient.

\begin{figure}[htb!]
    \centering
    \begin{subfigure}[t]{0.49\textwidth}
        \includegraphics[width=\textwidth]{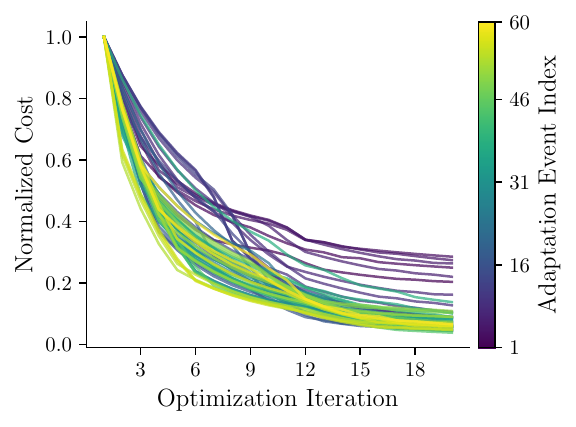}
        \vspace{-2em}
        \caption{{Case~1 (rich training)}}
    \end{subfigure}
    \hfill
    \begin{subfigure}[t]{0.49\textwidth}
        \includegraphics[width=\textwidth]{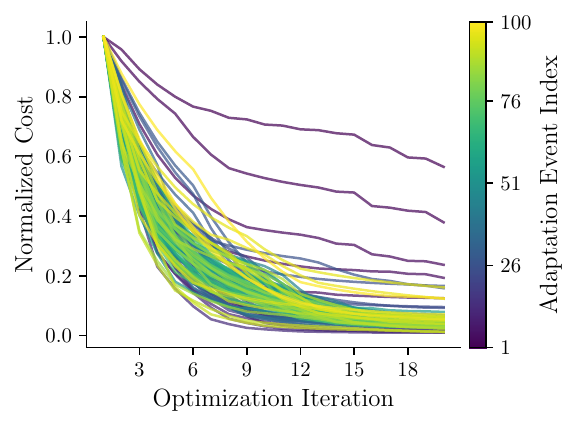}
        \vspace{-2em}
        \caption{{Case~2 (regime change)}}
    \end{subfigure}
    % \hspace{-1.em}
    \begin{subfigure}[t]{0.49\textwidth}
        \includegraphics[width=\textwidth]{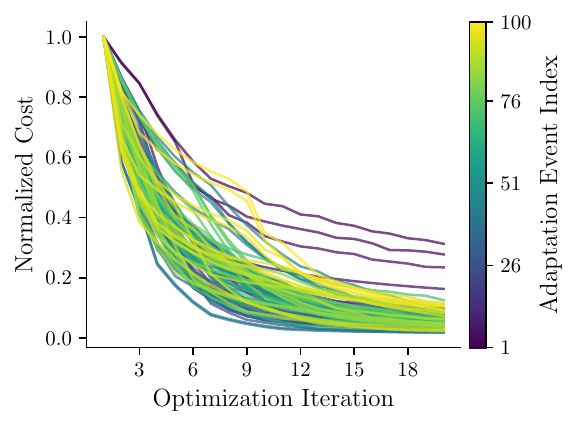}
        \vspace{-2em}
        \caption{{Case~3 (minimal training)}}
    \end{subfigure}
    
    \caption{\reviewerthree{Convergence histories of the online NiTROM optimization during adaptive updates for the
hybrid OpInf--NiTROM model. Each curve corresponds to a single adaptation event and shows
the normalized objective value over optimization iterations.}}
    \label{fig:nitrom_online_convergence}
\end{figure}

\paragraph{v. Convergence behavior of online NiTROM updates.}
\reviewerthree{For NiTROM-based adaptive ROMs, convergence analysis differs fundamentally from the
offline setting, as the optimization is re-initialized and solved repeatedly at each adaptation
event during online prediction. To assess the behavior of the online optimization, we examine
the convergence histories of the hybrid OpInf--NiTROM updates across all adaptation events.
Figure~\ref{fig:nitrom_online_convergence} shows the normalized objective value over optimization iterations for each adaptation
event in cases~1--3. While individual events exhibit different convergence rates, the objective
consistently decreases within a small number of iterations across all cases. This indicates
that a fixed, modest number of optimization steps is sufficient to obtain effective online
updates, without requiring event-specific convergence tuning.}

\paragraph{vi. Broader implications.}
Overall, the results demonstrate that online adaptation can significantly
extend the predictive range of non-intrusive ROMs.
When the system remains within or near the original subspace
(case 1), adaptive updates fully correct amplitude drifts.
When the dynamics evolve beyond that subspace (cases 2 and 3),
adaptation still provides partial recovery, especially
with hybrid updates.
Table~\ref{tab:summary} summarizes the qualitative performance trends
observed across all three cases.

\begin{table}[t]
  \centering
  \caption{Qualitative summary of ROM performance across cases.}
  \label{tab:summary}
  \begin{tabular}{lccc}
    \toprule
    \textbf{Model} & \textbf{Case 1} & \textbf{Case 2} & \textbf{Case 3} \\
    \midrule
    Static Galerkin/OpInf/NiTROM & Partially succeeds & Fails & Fails \\
    Adaptive OpInf        & Excellent robustness & Partially fails & Fails \\
    Adaptive NiTROM       & Accurate for small $Z$ & Fails & Fails \\
    Adaptive OpInf--NiTROM & \textbf{Best overall} & \textbf{Best overall} & \textbf{Best overall}  \\
    \bottomrule
  \end{tabular}
\end{table}

\paragraph{vii. A note on wall-clock time in Table~\ref{tab:wallclock}.}
Currently, these methods have a large overhead as we use unmodified offline versions of the SVD, OpInf and NiTROM constructions. There exist many opportunities to accelerate  ROMs by  a) updating algorithms to online, incremental contexts; and b) taking advantage of the fact that ROMs allow for larger time steps than FOMs.
We also remark that the cavity flow under consideration is not meant to particularly highlight computational savings. In fact, the small stencil width of the spatial discretization, the explicit time stepping, the relatively small $100\times 100$ grid, and Python vectorization, makes the FOM solver extremely fast (as demonstrated in the table).
More complicated problems requiring implicit time stepping, higher per-time-step cost (e.g., compressible flow solvers with shock-capturing logic), and much finer grids, would be better suited to demonstrate cost reduction.
As mentioned previously, however, a major objective of this work is to quantify challenges in achieving predictive model reduction. This particular problem was chosen because the FOM solver was very efficient (allowing us to run it hundreds of times), and because {\em even} this flow configuration exhibits challenging dynamical features for ROMs, and  allow us to identify promising research directions to ultimately enable the deployment of adaptive ROMs on transients- and advection-dominated flow problems.

The table is nonetheless useful to identify trends.
In particular, it becomes apparent that NiTROM-based adaptation is much more expensive than OpInf because the optimization requires ambient-space matrix-vector and matrix-matrix operations to optimize the test and trial subspaces against the available data.
Given the demonstrated benefits of the NiTROM refinement on the OpInf adaptive models, future work will focus on addressing this issue and reducing the cost associated with matrix-manifold optimization.

%%%%%%%%%%%%%%%%%%%%%%%%%%%%%%%%%%%%%%%%%%%%%%%%%%%%%%%%%%%%%
\section{Conclusion and Outlook}
\label{sec:conclusion}
%%%%%%%%%%%%%%%%%%%%%%%%%%%%%%%%%%%%%%%%%%%%%%%%%%%%%%%%%%%%%

This work introduced a general framework and a set of design rules to construct \emph{adaptive}, \emph{non-intrusive} reduced-order models (ROM), with the ultimate objective of extending traditional,
data-driven ROMs beyond the static setting and toward
continuously learning, streaming surrogates.  
The motivation is to design ROMs that not only compress and
approximate a complex dynamical system, but also \emph{evolve with it}.
To this end, we propose a  comprehensive formulation of
adaptive non-intrusive ROMs, together with three concrete realizations:
Adaptive Operator Inference (OpInf), Adaptive Non-intrusive Trajectory-based optimization of ROM (NiTROM), and a hybrid
Adaptive OpInf--NiTROM model that merges their respective advantages.  
These methods enable on-the-fly parameter correction using only
high-quality snapshots, without any intrusive access to the full-order
operators.

Systematic experiments on the two-dimensional lid-driven cavity flow
demonstrated that the proposed framework can substantially improve
long-term stability and accuracy compared with
conventional static ROMs.  
When the model operates near its training regime, adaptive
updates effectively eliminate energy drift and preserve amplitude and phase alignment
over long horizons.  
Under more difficult extrapolation settings, where the system transitions
into previously unseen regimes, the hybrid model
consistently yields the most physically meaningful and stable results.
Together, these findings establish the feasibility of truly
\emph{self-correcting} non-intrusive ROMs that can adapt to evolving
dynamics in real time.

Despite these promising outcomes, several  limitations remain.
While the NiTROM-based models demonstrate high accuracy in
moderate adaptation cases, their manifold optimization steps are computationally
intensive and sensitive to hyperparameters such as the adaptation
window size, lookback window size, and the number of online optimization steps.
Although these methods are  less expensive than full-order simulations,
their current cost and sensitivity make deployment
challenging in digital twin-type settings.  We are also careful to acknowledge that rigorous testing and evaluations are required on a broader range of test problems and data scenarios.
At present, the framework represents a \emph{proof of concept} or solver accelerator. 
Addressing these limitations will require continued algorithmic
development to improve robustness and efficiency.

\noindent {\bf Outlook: }
Several directions naturally emerge for future research. Firstly, as presented, the maximum speedup of the proposed algorithms is strictly upper bounded by the frequency of adaptation $Z$, as the full high-fidelity operator is evolved  every $Z$ time-steps. This may be accelerated by a sampling of the high fidelity operator followed by interpolation, as in classical intrusive ROMs.
Second, developing adaptive, error-based trigger mechanisms could
replace the fixed adaptation window with intelligent decisions driven
by residual indicators.
Third, acceleration of the Adaptive OpInf formulation through
incremental or low-rank operator updates could move the method closer to
real-time feasibility, especially for high-frequency adaptation. 
Finally, testing the framework in a true digital twin environment, where
adaptation is informed directly by sensor data rather than full-order
feedback, represents an essential step toward real-world applications.

Overall, this study provides the first detailed exploration of adaptive
non-intrusive reduced-order modeling and demonstrates that continuous,
data-driven adaptation is necessary to train \emph{truly} predictive models. The results establish a foundation for the next generation of self-correcting, low-dimensional surrogates capable of long-term, physics-consistent prediction in complex, non-stationary systems.
We anticipate that such adaptive ROMs will play a
central role in future developments in digital twins,
model-based control, and scientific machine learning for
high-dimensional dynamical systems.

%%%%%%%%%%%%%%%%%%%%%%%%%%%%%%%%%%%%%%%%%%%%%%%%%%%%%%%%%%%%%
\begin{appendices}
%%%%%%%%%%%%%%%%%%%%%%%%%%%%%%%%%%%%%%%%%%%%%%%%%%%%%%%%%%%%%

\section{Additional flowfield visualizations}
\label{app:fields}

\begin{figure}[t]
  \centering
  \includegraphics[width=\linewidth]{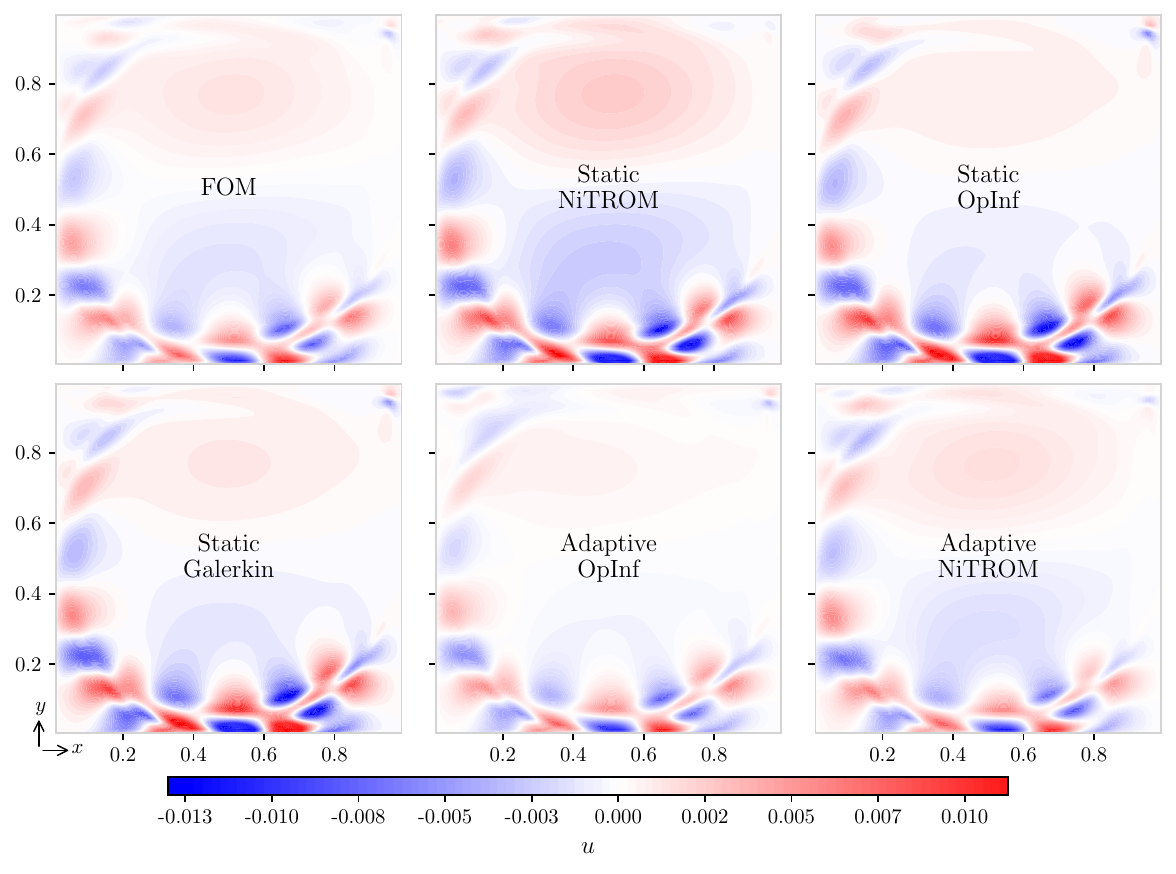}
  \caption{Ground truth (FOM) and predicted (ROM) $u$-velocity fields at the end of the online window ($t=20$) in case 1. All models are quadratic and $r=10$-dimensional. For adaptive models we have an adaptation window $Z=10$, a lookback window $M=100$, and a per-adaptation optimization budget $K=10$.}
  \label{fig:case1_u}
\end{figure}

\begin{figure}[t]
  \centering
  \includegraphics[width=\linewidth]{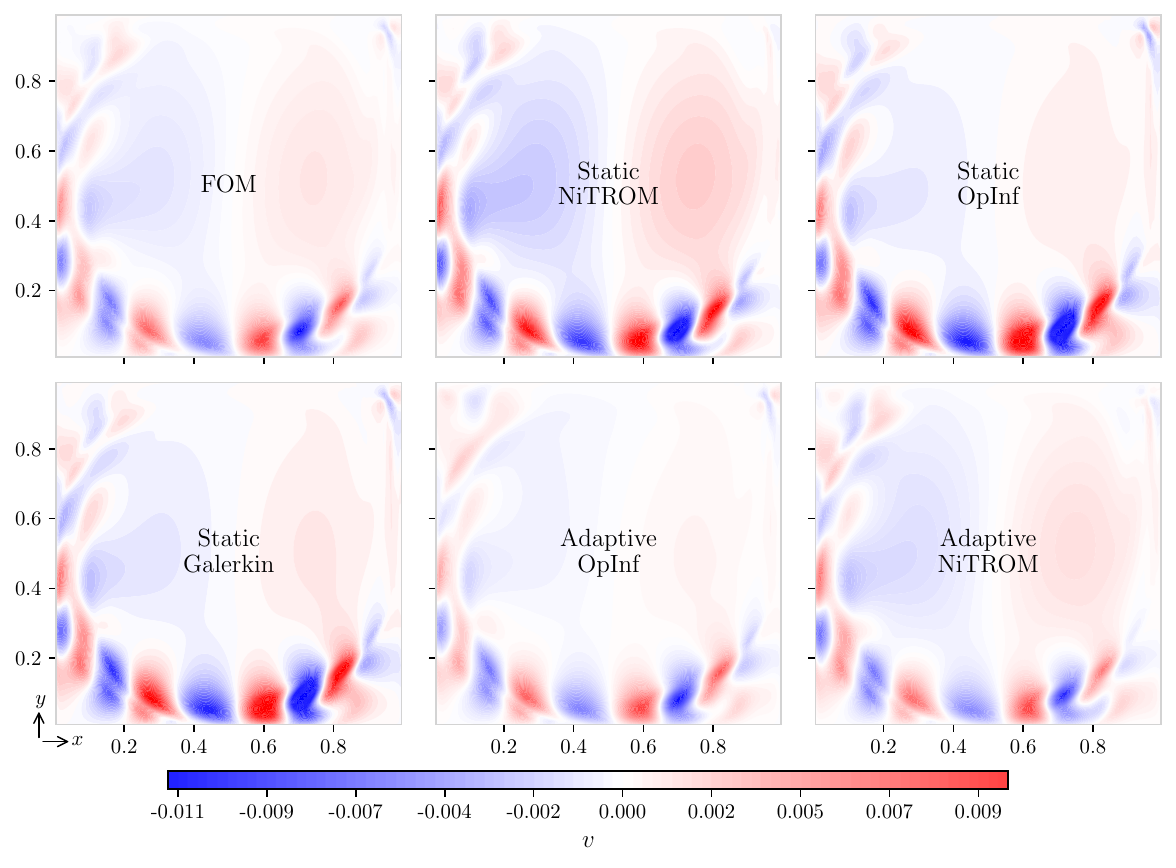}
  \caption{Ground truth (FOM) and predicted (ROM) $v$-velocity fields at the end of the online window ($t=20$) in case 1. All models are quadratic and $r=10$-dimensional. For adaptive models we have an adaptation window $Z=10$, a lookback window $M=100$, and a per-adaptation optimization budget $K=10$.}
  \label{fig:case1_v}
\end{figure}

\begin{figure}[t]
  \centering
  \includegraphics[width=\linewidth]{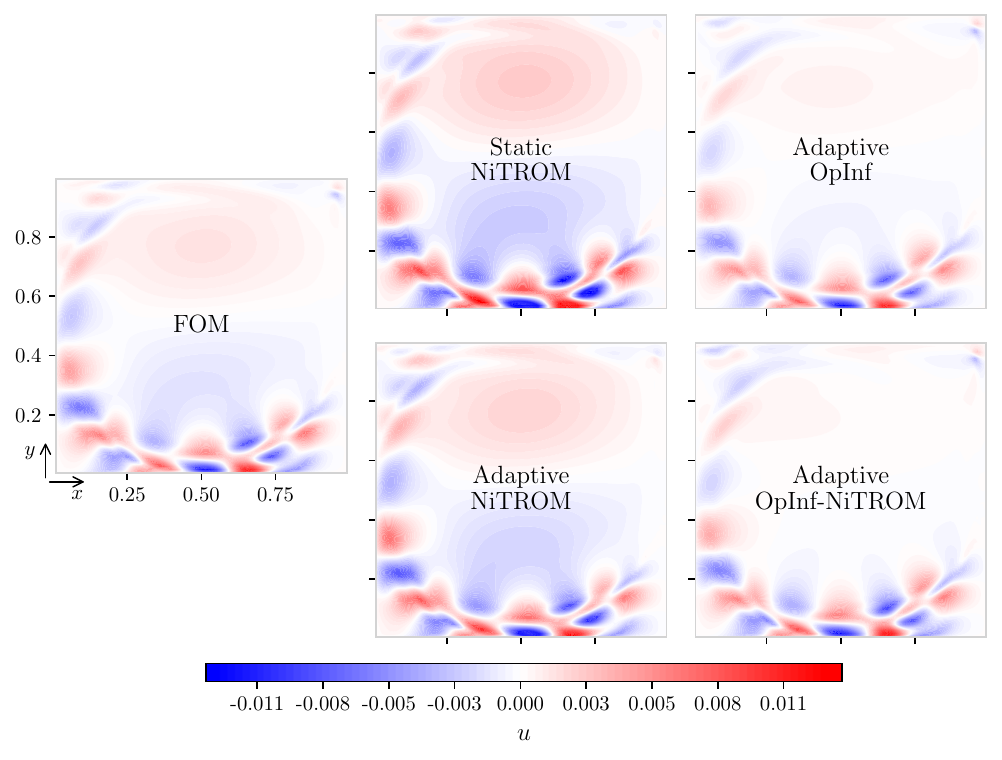}
  \caption{Ground truth (FOM) and predicted (ROM) $u$-velocity fields at the end of the online window ($t=20$) in case 1. All models are quadratic and $r=10$-dimensional. For adaptive models we have an adaptation window $Z=50$, a lookback window $M=20$, and a per-adaptation optimization budget $K=10$.}
  \label{fig:case1_z50_u}
\end{figure}

\begin{figure}[t]
  \centering
  \includegraphics[width=\linewidth]{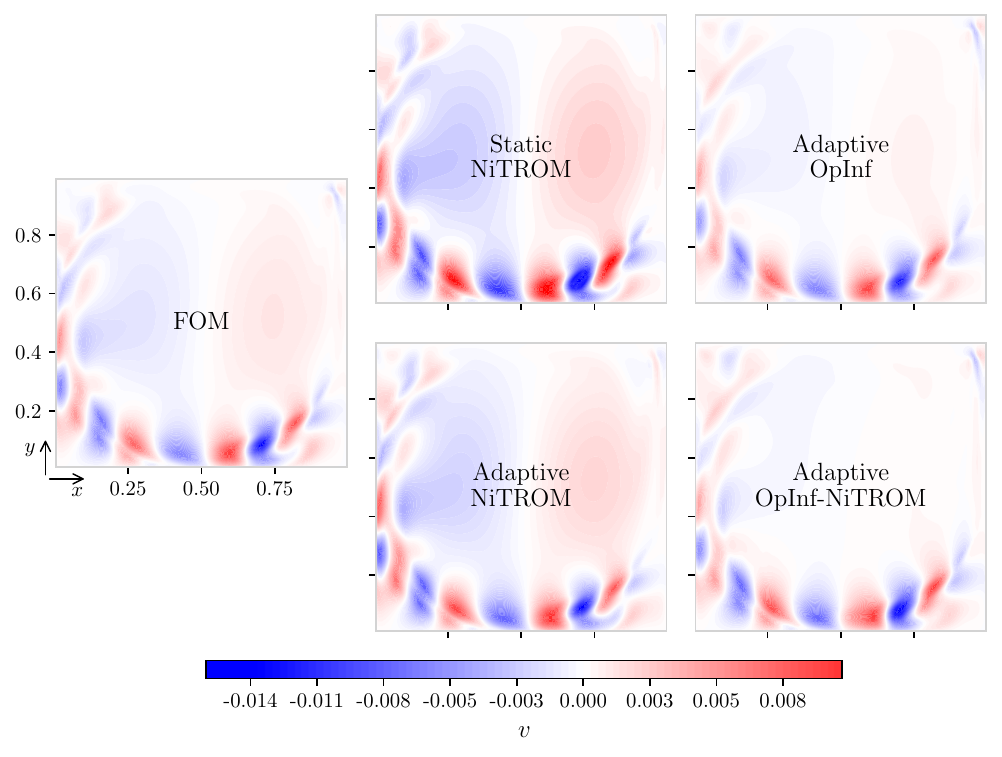}
  \caption{Ground truth (FOM) and predicted (ROM) $v$-velocity fields at the end of the online window ($t=20$) in case 1. All models are quadratic and $r=10$-dimensional. For adaptive models we have an adaptation window $Z=50$, a lookback window $M=20$, and a per-adaptation optimization budget $K=10$.}
  \label{fig:case1_z50_v}
\end{figure}

\begin{figure}[t]
  \centering
  \includegraphics[width=\linewidth]{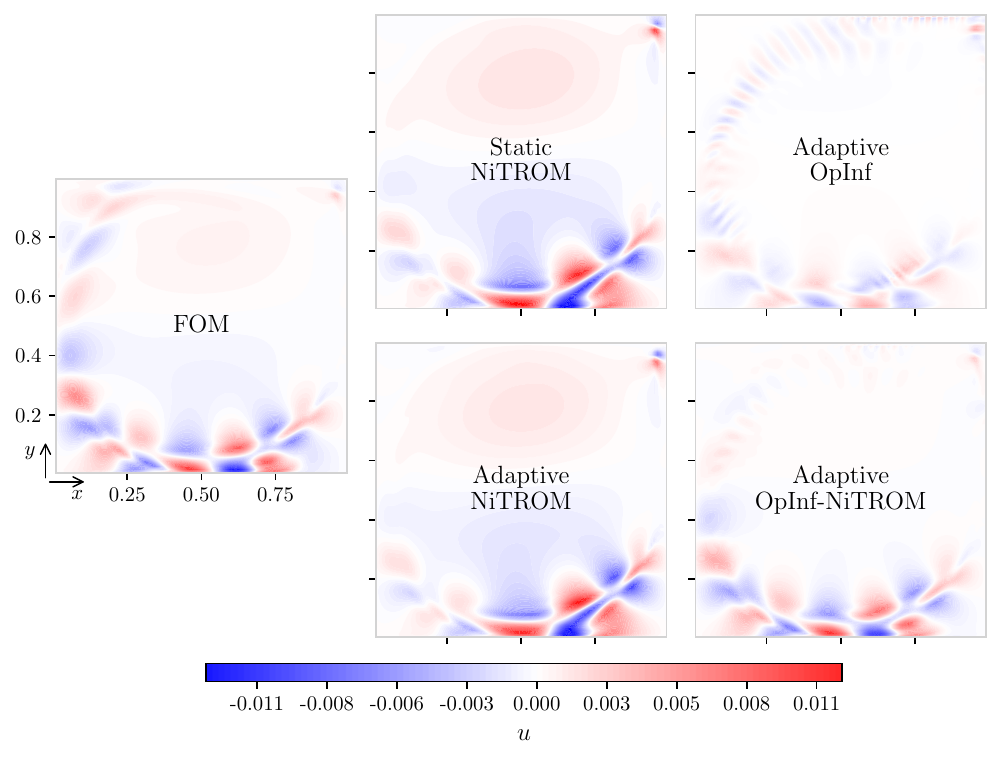}
  \caption{Ground truth (FOM) and predicted (ROM) $u$-velocity fields at the end of the online window ($t=10$) in case 2. All models are quadratic and $r=10$-dimensional. For adaptive models we have an adaptation window $Z=10$, a lookback window $M=10$, and a per-adaptation optimization budget $K=10$.}
  \label{fig:case2_u}
\end{figure}

\begin{figure}[t]
  \centering
  \includegraphics[width=\linewidth]{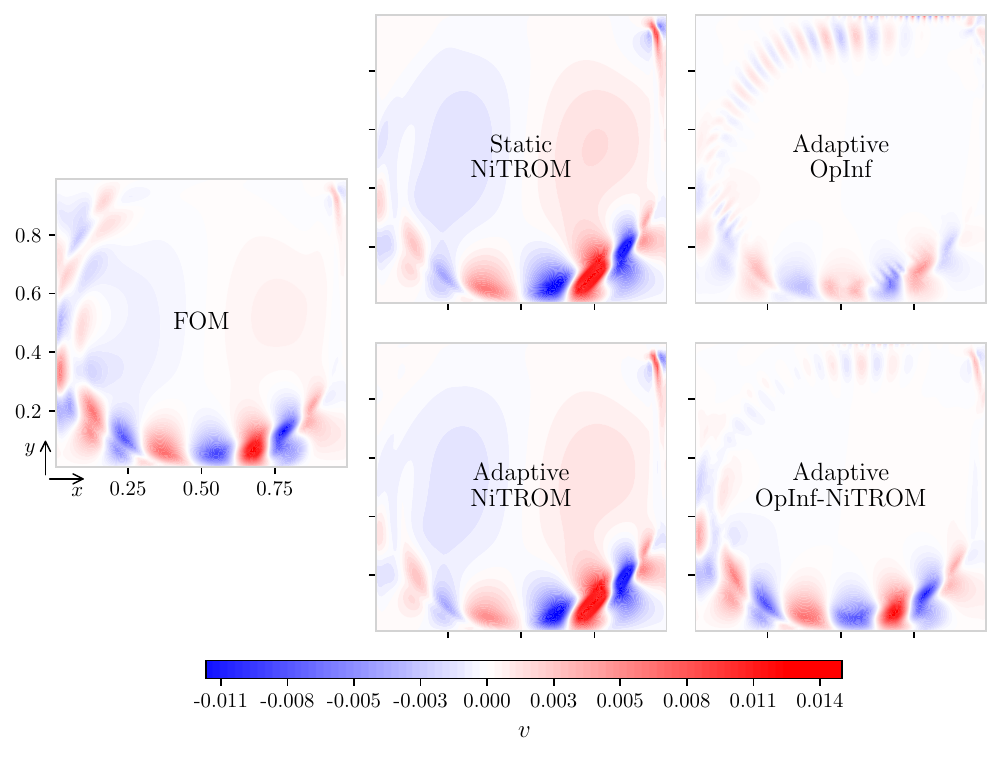}
  \caption{Ground truth (FOM) and predicted (ROM) $v$-velocity fields at the end of the online window ($t=10$) in case 2. All models are quadratic and $r=10$-dimensional. For adaptive models we have an adaptation window $Z=10$, a lookback window $M=10$, and a per-adaptation optimization budget $K=10$.}
  \label{fig:case2_v}
\end{figure}

\begin{figure}[t]
  \centering
  \includegraphics[width=\linewidth]{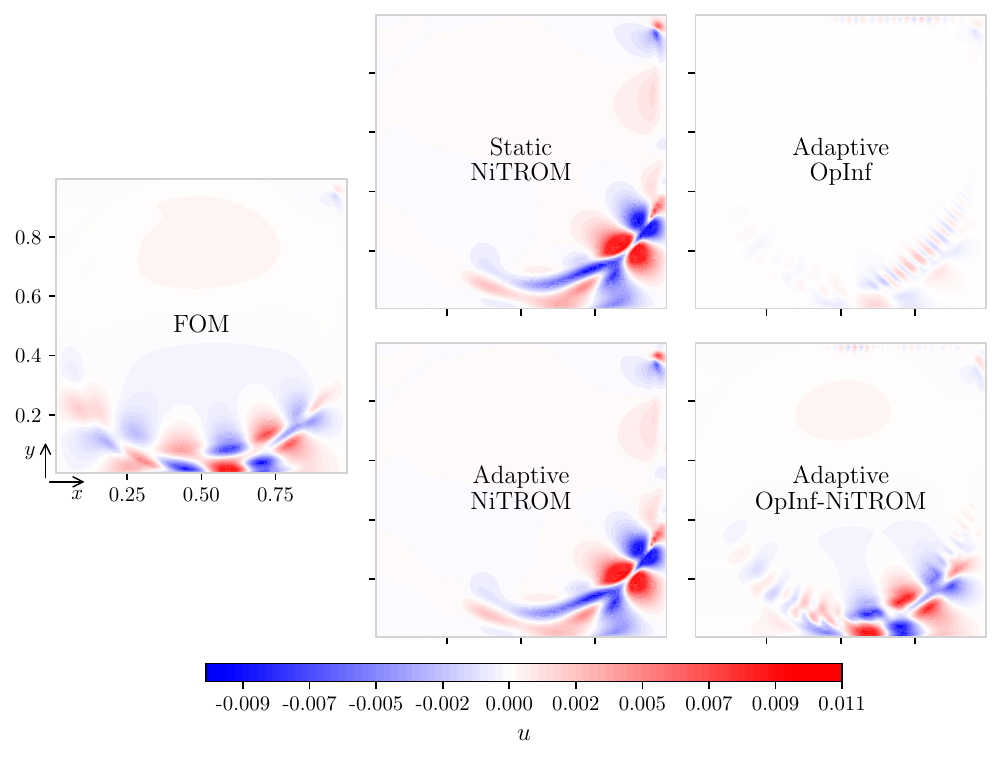}
  \caption{Ground truth (FOM) and predicted (ROM) $u$-velocity fields at the end of the online window ($t=6.25$) in case 3. All models are quadratic and $r=10$-dimensional. For adaptive models we have an adaptation window $Z=10$, a lookback window $M=10$, and a per-adaptation optimization budget $K=10$.}
  \label{fig:case3_u}
\end{figure}

\begin{figure}[t]
  \centering
  \includegraphics[width=\linewidth]{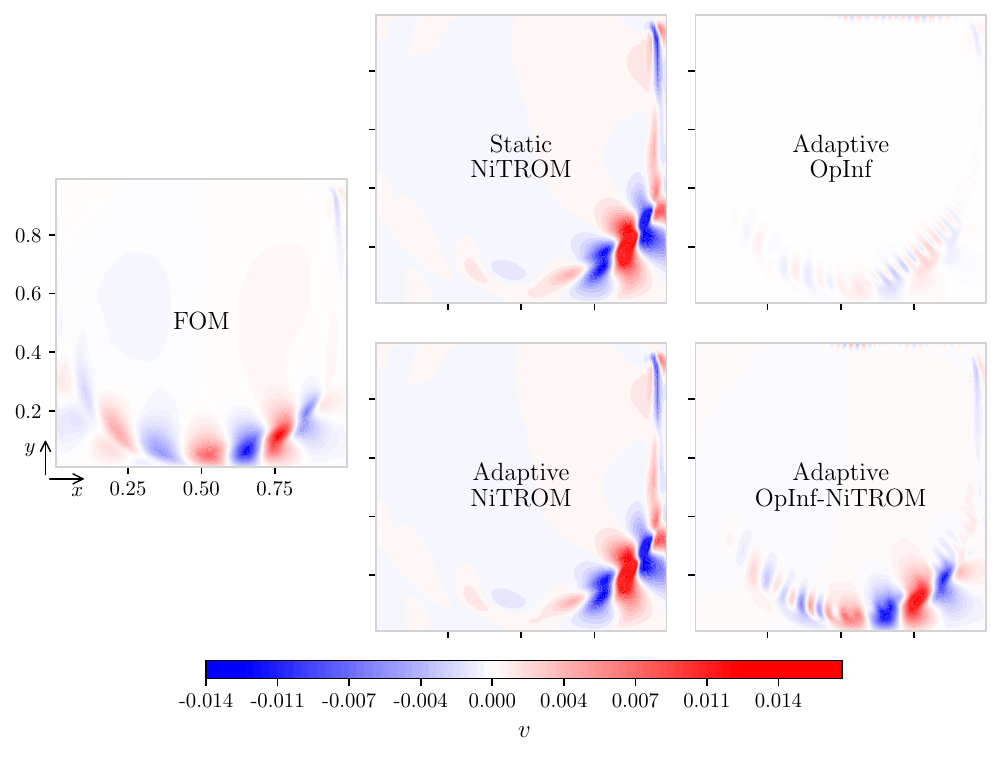}
  \caption{Ground truth (FOM) and predicted (ROM) $v$-velocity fields at the end of the online window ($t=6.25$) in case 3. All models are quadratic and $r=10$-dimensional. For adaptive models we have an adaptation window $Z=10$, a lookback window $M=10$, and a per-adaptation optimization budget $K=10$.}
  \label{fig:case3_v}
\end{figure}

To complement the vorticity visualizations in Sec.~\ref{sec:results}, this appendix provides the corresponding $u$- and $v$-velocity fields for all cases (Figs.~\ref{fig:case1_u}--\ref{fig:case3_v}). These figures illustrate the same qualitative trends discussed in Sec.~\ref{sec:results}: all adaptive models do well in suppressing the non-physical energy growth in case 1, while the hybrid model is the most effective approach in more challenging settings (cases 2 and 3).

\end{appendices}

%%%%%%%%%%%%%%%%%%%%%%%%%%%%%%%%%%%%%%%%%%%%%%%%%%%%%%%%%%%%%
\section*{Statements and Declarations}

\subsection*{Funding}
A.H. and K.D. were supported by the AFOSR grant \#FA9550-17-10195 and OUSD(RE) Grant \#N00014-21-1-295. A.P.  acknowledges startup support from the Newark College of Engineering at the New Jersey Institute of Technology.

\subsection*{Author contribution}
{\bf A.H.} contributed to methodology, software development, numerical experiments, and writing---original draft and review \& editing.
{\bf A.P.} contributed to methodology, software development, supervision, and writing---review \& editing.
{\bf K.D.} contributed to conceptualization, funding acquisition, resources, supervision, and writing---review \& editing.

\subsection*{Replication of results (data availability)}
The source code and data required to reproduce the results will be made available upon request to the corresponding author.

\subsection*{Conflict of interest}
On behalf of all authors, the corresponding author declares that there are no conflicts of interest associated with this work.

\subsection*{Ethics approval and consent to participate}
Not applicable.

\subsection*{Acknowledgments}
The authors acknowledge insightful discussions with Prof. Laura Balzano (University of Michigan).

\bibliography{bibliography}

\end{document}